\documentclass[10pt,twocolumn,letterpaper]{article}

\usepackage{cvpr}
\usepackage{times}
\usepackage{epsfig}
\usepackage{graphicx}
\usepackage{amsmath}
\usepackage{amssymb}

\usepackage{floatrow}
\usepackage{url}
\usepackage{color}
\usepackage{xcolor}
\usepackage{empheq}
\usepackage{amsfonts}
\usepackage{dsfont}
\usepackage{caption}
\usepackage{subcaption}
\usepackage{enumitem}
\usepackage{multirow}
\usepackage{booktabs}
\usepackage{threeparttable}
\usepackage{wrapfig}
\usepackage{makecell}
\usepackage[]{algorithm2e}
\usepackage{tablefootnote}
\usepackage{footnote}
\usepackage{amsthm}

\usepackage[pagebackref=true,breaklinks=true,letterpaper=true,colorlinks,bookmarks=false]{hyperref}

\newfloatcommand{capbtabbox}{table}[][\FBwidth]
\makesavenoteenv{tabular}
\makesavenoteenv{table}

\newcommand*{\tran}{{^{\mkern-1.5mu\mathsf{T}}}}
\newcommand{\w} {{ \mathbf{w} }}
\newcommand{\x} {{ \mathbf{x} }}
\newcommand{\y} {{ \mathbf{y} }}
\newcommand{\BN} {{ \mathrm{bn} }}
\newcommand{\IN} {{ \mathrm{in} }}
\newcommand{\LN} {{ \mathrm{ln} }}
\newcommand{\WN} {{ \mathrm{wn} }}

\cvprfinalcopy % *** Uncomment this line for the final submission

\ifcvprfinal\fi
\begin{document}

%%%%%%%%% TITLE
\title{Do Normalization Layers in a Deep ConvNet Really Need to Be Distinct?\thanks{{\small This is a preprint version. The authors share contributions. Corresponding to: Ping Luo $<$pluo.lhi@gmail.com$>$, Zhanglin Peng $<$pengzhanglin@sensetime.com$>$, Jiamin Ren $<$renjiamin@sensetime.com$>$, Ruimao Zhang $<$ruimao.zhang@cuhk.edu.hk$>$.}}}

\author{Ping Luo$^\dagger$\hspace{30pt}Zhanglin Peng$^\ddagger$\hspace{30pt}Jiamin Ren$^\ddagger$\hspace{30pt}Ruimao Zhang$^\dagger$\\
$^\dagger$The Chinese University of Hong Kong\hspace{30pt}$^\ddagger$SenseTime Research\\
}

\maketitle
%\thispagestyle{empty}

%%%%%%%%% ABSTRACT
\begin{abstract}
    Yes, they do. This work investigates a perspective for deep learning: whether different normalization layers in a ConvNet require different normalizers.
    This is the first step towards understanding this phenomenon. We allow each convolutional layer to be stacked before a
    switchable normalization (SN) that learns to choose a normalizer from a pool of normalization methods.
   Through systematic experiments in ImageNet, COCO, Cityscapes, and ADE20K, we answer three questions:
   (a) Is it useful to allow each normalization layer to select its own normalizer?
   (b) What impacts the choices of normalizers?
   (c) Do different tasks and datasets prefer different normalizers?
   Our results suggest that (1) using distinct normalizers improves both learning and generalization of a ConvNet; (2) the choices of normalizers are more related to depth and batch size, but less relevant to parameter initialization, learning rate decay, and solver;
   (3) different tasks and datasets have different behaviors when learning to select normalizers.
\end{abstract}

%%%%%%%%% BODY TEXT
\section{Introduction}

The successes of deep learning in computer vision can be attributed to the increase of two factors: (1) data size and (2) depth of network.
For example, there are often hundreds of layers in the recent convolutional neural networks (CNNs), including hand-crafted architectures such as Inception \cite{C:Inception-v1,C:Inception-v3}, ResNet \cite{C:resnet} and DenseNet \cite{C:DenseNet}, device-friendly architectures such as MobileNet \cite{MobileNets} and ShuffleNet \cite{ShuffleNet}, as well as neural architecture searches \cite{DARTS,ENAS}.

The above CNNs typically stacked a basic network many times to build deep models.
%,
This ``atomic'' network consists of a convolutional layer, a normalization layer, and an activation function.
It can be seen that the normalization method is an indispensable component in these deep models, where each model may have tens of normalization layers.
However, existing CNNs assumed that all normalization layers use the same normalization approach uniformly such as batch normalization (BN) \cite{C:BN}, resulting in non-optimal performance.

\textbf{Overview.} This work presents the first systematical study of a novel perspective in deep learning: \emph{whether different convolutional layers in a CNN should use different normalizers}. This viewpoint would impact many vision problems.
We employ Switchable Normalization (SN) \cite{SN} as our approach, which is a recently-proposed normalization method that
learns to select appropriate normalizer such as BN, instance normalization (IN) \cite{IN}, and layer normalization (LN) \cite{LN} for each normalization layer of a CNN.

We first explain necessary mathematics of our empirical setups and then investigate the selectivity of normalizers in three important vision tasks, including image recognition in ImageNet \cite{C:ImageNet}, object detection in COCO \cite{lin2014microsoft}, and scene segmentation in Cityscapes \cite{C:cityscape} and ADE20K \cite{A:ADE20K}. The `selectivity' is defined as the learning dynamics when selecting normalizers. This work answers the following three questions.

\begin{figure*}[t]
  \centering
  \includegraphics[width=0.9\linewidth]{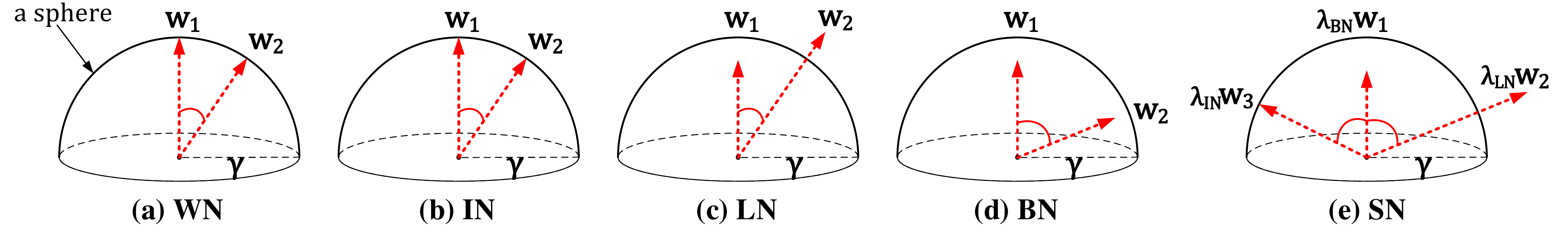}
  \caption{{\small Geometric comparisons of WN, IN, LN, BN, and SN. These normalizers are compared in an unify way by represent them by using WN that decomposes optimization of filters into their directions and lengths. In this way, IN is identical to WN that sets the filter norm to `1' (\ie $\|\w_1\|=\|\w_2\|=1$) and then rescales them to $\gamma$. LN is less constrained than IN and WN to increase learning ability. BN increases angle between filters and reduces filter length to improve generalization. SN inherits all their benefits by learning their importance ratios.
  }}\label{fig:SN}
\end{figure*}

\begin{itemize}
	\item \emph{Is it useful to allow each normalization layer to use its own normalization operation?} We show that performance in ImageNet can be improved by placing distinct normalizers in appropriate positions of a network, because they have different properties that would help learning image representation.
The learned features and chosen normalizers are transferable to COCO, Cityscapes, and ADE20K.

\item \emph{What impacts the choices of normalizers?}
   By studying their learning dynamics in ImageNet,
   we find that the selection is more sensitive to depth, batch size, and input image size, but less relevant to random parameter initialization, learning rate decay (\eg stepwise decay, cosine decay \cite{A:SGDR}), and solver (\eg SGD+momentum, RMSProp).

\item \emph{Do different networks, tasks, and datasets prefer different normalizers?} Our empirical results suggest that they prefer different dynamics and configurations of normalizers.
\end{itemize}

We make three key \textbf{contributions}. (1) This is the first work to investigate the impacts of using different normalization methods after different convolutional layers within a CNN. We verify this viewpoint in computer vision.
(2) We identify key factors that affect the selectivity of normalizers.
Our findings are useful in many vision tasks and may inspire the other problems that are not presented in this work.
(3) We make this study completely reproducible by organizing a codebase that contains the pretrained or finetuned models in many benchmarks.
This codebase will be released.

In remaining sections, we introduce approaches and related work in Sec.\ref{sec:SN}.
Sec.\ref{sec:train} discusses pretraining and finetuning methods. Sec.\ref{sec:exp} presents experiments.

\section{Approaches and Setups}\label{sec:SN}

This section explains our approaches and empirical setups.

\textbf{Selecting Normalizers.} To investigate the selectivity of normalizers, we adopt switchable normalization (SN) \cite{SN} that learns to choose a normalizer from a set of normalization methods. SN is applied after each convolutional layer,
\begin{equation}\label{eq:SN}
\vspace{-5pt}
\hat{h}_{\mathrm{sn}}=\frac{h-\Sigma_{z\in\Omega} \lambda_z^\mu\mu_z}
{{\Sigma_{z\in\Omega}\lambda_z^\sigma\sigma_z}},
\end{equation}
where $h$ and $\hat{h}_{\mathrm{sn}}$ are pixel values of each hidden channel before and after normalization, $\mu_z=\mathds{E}[h],\sigma_z=\sqrt{\mathds{E}[h^2]-\mathds{E}[h]^2}$ are mean and standard deviation of $h$ estimated by using a certain normalizer $z$, and $\Omega$ indicates a set of normalizers. $\lambda_z^\mu$ and $\lambda_z^\sigma$ are importance ratios of mean and variance respectively.
They are learnable parameters by using softmax function. We have $\forall \lambda_z^\mu,\lambda_z^\sigma\in[0,1]$, $\Sigma_{\forall z\in\Omega}\lambda_z^\mu=1$ and $\Sigma_{\forall z\in\Omega}\lambda_z^\sigma=1$.
Existing work also applied a linear transformation after $\hat{h}$ and before ReLU, by learning a scale parameter $\gamma$ and a bias parameter $\beta$, that is $\gamma\hat{h}_{\mathrm{sn}}+\beta$.
Eqn.\eqref{eq:SN} shows that each pixel is normalized by using a weighted average of statistics, which are estimated by a set of normalizers in $\Omega$.

\textbf{$\boldsymbol{\lambda_z^\mu}$ and $\boldsymbol{\lambda_z^\sigma}$.}
We point out that the important ratios $\lambda_z^\mu$ and $\lambda_z^\sigma$ are not identical in Eqn.\eqref{eq:SN}.
We are interested in this setup because ${\mu_z}$ and ${\sigma_z}$ have different impacts in training.
We take BN as an example to understand this. The impacts of ${\mu_\BN}$ and ${\sigma_\BN}$ can be distinguished by using weight normalization (WN) \cite{WN}, which is written by $\hat{h}_{\mathrm{wn}}=\frac{h}{\|\w\|_2}$ and $h=\w\tran\x$ where $\w$ and $\x$ represent a filter and an image patch respectively.
Note that BN is reduced to WN by assuming ${\mu_\BN}=0$ and $\sigma_\BN=1$ \cite{WN}.
In fact, a network trained with WN might not perform as well as BN.
However, by investigating how well mean-only BN and std-only BN\footnote{WN trained with mean-only BN or std-only BN are achieved by stacking a BN layer after a WN layer, where either $\mu$ or $\sigma$ are used in BN.} would improve WN, we can distinguish the impacts from ${\mu_\BN}$ and ${\sigma_\BN}$.

In fact, previous work \cite{WN,regBN} found that WN trained with either mean-only or std-only BN cannot outperform BN, and both of them are important to achieve comparable performance to BN.
% but they have different improvements.
%
Therefore, we treat ${\lambda_z^\mu}$ and ${\lambda_z^\sigma}$ differently throughout our studies to better understand their different behaviors.

\textbf{Properties of Normalizers in $\boldsymbol{\Omega}$.}
We define $\Omega=\{\mathrm{IN},\mathrm{LN},\mathrm{BN}\}$.
We are not going to exhaustively enumerate the methods in $\Omega$, where
%,
BN, IN, and LN are chosen as representatives. Their importance ratios in training are sufficient to show the learning behaviors of distinct normalizers.
To see this, we can also represent IN and LN by using WN. Therefore, the characteristics of BN, IN, LN, and SN can be compared in a unify way.

Specifically, the computation of WN is defined by $\hat{h}_\WN=\frac{h}{\|\w_i\|_2}=\frac{\w_i\tran\x}{\|\w_i\|_2}$, which normalizes the norm of each filter to `1'. We use $i$ to indicate a filter of the $i$-th channel.
As shown in Fig.\ref{fig:SN}(a), WN normalizes the norm of each filter to a unit sphere with length `1', and then rescales the length to $\gamma$ that is a learnable scale parameter. To simplify our discussions, we suppose this scale parameter is shared among all channels.
In other words, WN decomposes the optimization of a filter into its direction and length.

By assuming $\mu_z=0$ and $\sigma_z=1$ for all the normalizers in $\Omega$, we see that they can be also represented by using WN.
For instance, IN turns into $\hat{h}_\IN=\frac{h-\w_i\tran\mathds{E}[\x]}{\sqrt{\w_i\tran(\mathds{E}[\x\x\tran]
-\mathds{E}[\x]\tran\mathds{E}[\x])\w_i}}=\frac{\w_i\tran\x}{\|\w_i\|_2}$ that is identical to WN as shown in Fig.\ref{fig:SN}(b).
For LN, each channel is normalized by all the channels and thus $\hat{h}_\LN=\frac{\w_i\tran\x}{\sqrt{\frac{1}{C}\sum_{i=1}^C\|\w_i\|_2^2}}$ where $C$ is the number of channels.
The filter norm in LN is less constrained than WN and IN as visualized in Fig.\ref{fig:SN}(c), where the filter norm can be either longer or shorter than $\gamma$, in the sense that LN increases learning capacity of the networks by diminishing regularization.

Furthermore, BN can be represented by WN as discussed before.
Luo \etal \cite{regBN} showed that BN imposes regularization on norms of all the filters and reduces correlations between filters.
Its geometry interpretation can be viewed in Fig.\ref{fig:SN}(d), where the filter norms would be shorter and angle between filters would be larger than the other normalizers.
In conclusion, BN improves generalization \cite{C:BN}. In general, we would have the following relationships
\vspace{-5pt}
\begin{eqnarray}\label{eq:order}
&\mathrm{learning~capacity}: LN>BN>IN,\nonumber\\
&\mathrm{generalization~capacity}: BN>IN>LN. \nonumber
\end{eqnarray}
\vspace{-15pt}

Finally, SN can also be interpreted by using WN.
To see this, we consider a simple case when $\lambda_z^\mu=\lambda_z^\sigma$. In this case, Eqn.\eqref{eq:SN} can be reformulated as $\hat{h}_{\mathrm{sn}}=\lambda_\IN\hat{h}_{\IN}+\lambda_\LN\hat{h}_{\LN}+\lambda_\BN\hat{h}_{\BN}$ by combing the above normalizers.
It is seen that SN aggregates benefits from all three approaches as shown in Fig.\ref{fig:SN}(e), by learning their important ratios to adjust both learning and generalization capacity.

\subsection{Connections with Previous Work}\label{sec:con}

\textbf{Normalization Methods.} There are normalization approaches in the literature other than those mentioned above. Although they are not considered in $\Omega$, we would like to acknowledge their contributions. For example, group normalization (GN) \cite{GN} divides channels into groups. Let $g$ be the number of groups. GN can be represented by WN as $\hat{h}_{\mathrm{gn}}=\frac{\w_i\tran\x}{\sqrt{\frac{g}{C}\sum_{i=1}^{{C/g}}\|\w_i\|_2^2}}$ that is a special case of LN. Since GN introduces an additional hyper-parameter and its learning behavior would be similar to LN, we do not add GN in $\Omega$.
Furthermore, batch renormalization (BRN) \cite{BRN} and batch kalman normalization (BKN) \cite{BKN} extended BN to account for training in small batch size.
And divisive normalization (DN) \cite{Normalizing} normalized each pixel by using its neighboring region.

\textbf{Understanding Depth.} This work investigates depth of CNN with respect to the selectivity of normalizers.
There are also many studies that explored depth of network from the other aspects.
We review some representatives.
Ba \etal \cite{DoFC} showed that by using a shallow student network to mimic a deep teacher network, multilayer perceptron (MLP) can learn complex functions previously learned by deep CNNs while using similar number of parameters. This was the first study to show that MLPs are not necessary to be deep to achieve good performance.

Urban \etal \cite{DoCNN} found that the above observation does not apply when the student is a CNN. Many convolutional layers are required to learn functions of a deep teacher.
This study showed that a CNN student should be sufficiently deep to achieve good performance, although it may not be as deep as its teacher.
Moreover, Zhang \etal \cite{rethinking} showed that small generalization error of a deep network may not attribute to its depth and regularization techniques used in training, because we are far away from understanding
why CNNs are relatively easy to optimize and why they have good generalization ability.

The above studies shed light on foundations to understand deep learning, which might not have direct guidance in practice.
However, unlike them, our study explores a new viewpoint to understand and to design deep models, which is of great value in many practical problems.

\section{Training Methods}\label{sec:train}

Now we introduce pretraining and finetuning procedures that are used throughout this work.

\textbf{Pretraining.} For a CNN with a set of network parameters $\Theta$
and a set of training samples and their labels denoted as $\{\x_j,\y_j\}_{j=1}^N$, the training problem is formulated as minimizing an empirical loss function $\frac{1}{N}\sum_{j=1}^N\mathcal{L}(\y_j,f(\x_j;\Theta))$ with respect to $\Theta$, where $f(\x_j;\Theta)$ is a function learned by the CNN to predict $\y_j$.
When a CNN is trained with SN, we replace all its previous normalization layers such as BN by using SN.
This step introduces a set of control parameters ${\Phi}=\{\lambda_\mathrm{in}^{\mu\ell},\lambda_\mathrm{ln}^{\mu\ell},\lambda_\mathrm{bn}^{\mu\ell},
\lambda_\mathrm{in}^{\sigma\ell},\lambda_\mathrm{ln}^{\sigma\ell},\lambda_\mathrm{bn}^{\sigma\ell}\}_{\ell=1}^M$ for $M$ normalization layers.
Therefore, the optimization problem becomes $\min_{\{\Theta,\Phi\}}\frac{1}{N}\sum_{j=1}^N\mathcal{L}(\y_j,f(\x_j;\Theta,\Phi))$.
For example, when pretraining in ImageNet, both $\Theta$ and $\Phi$ are optimized jointly in a single feed-forward step by using SGD with momentum 0.9 and an initial learning rate 0.1, which is divided by 10 after 30, 60, and 90 epochs.

The above learning problem has important differences compared to meta-learning \cite{bilevel,hyperparameter,ENAS}, which are also able to learn control parameters. We move discussions to Appendix \ref{appsec:ML} due to length of the paper.

\textbf{Initializing SN.}
Given a CNN model, we initialize its network parameters, batch size, learning rate, and the other hyper-parameters (such as weight decay) by strictly following existing settings.
The only difference is that the normalization layer is replaced by SN. We initialize the scale parameter $\gamma=1$, the shift parameter $\beta=0$, and all the control parameters $\Phi=\frac{1}{3}$.
We also add weight decay of $10^{-4}$ to all these parameters.
We would like to point out that initializing $\gamma$ specially in certain network may improve performance. For example, Goyal \etal \cite{large-minibatch} showed that performance of ResNet50 \cite{C:resnet} would be improved when $\gamma$ of each residual block's last normalization layer is initialized as 0 but not 1.
%,
%
However, we didn't adopt any trick in order to keep parameter initializations as simple as possible.

\textbf{Finetuning.} We finetune a pretrained deep model by following its original protocols in the corresponding benchmarks.
BN in SN are not frozen and not synchronized across GPUs.
Moreover, BN in SN is evaluated by using batch average following \cite{SN} rather than moving average.

\section{Experiments}\label{sec:exp}

This section systematically investigates the selectivity behaviors in ImageNet \cite{C:ImageNet}, COCO \cite{lin2014microsoft}, Cityscapes \cite{C:cityscape}, and ADE20K \cite{A:ADE20K}.
In each benchmark, all models are trained by following common settings.
Details are moved to Appendix \ref{app:exp_proto}. By default, the models of SN are trained by following Sec.\ref{sec:train}.

\subsection{Image Recognition in ImageNet}\label{sec:img}

We first present results in ImageNet.

\textbf{Comparisons.} Table \ref{tab:comp} shows that ResNet50 trained with distinct normalizers (using SN)
outperforms when it is trained with BN, GN, IN, and LN alone. For example, SN surpasses BN by $0.7\%$ and $1.5\%$ in $(8,32)$ and $(8,8)$.
In small minibatch $(8,2)$, SN achieves the second-best performance next to GN.
It is seen that GN has uniform accuracy around $75.9$ for all batch sizes \cite{GN}, because it is independent with batch statistics. The accuracy of SN increases when batch size increases.
In Table \ref{tab:comp}, SN$_{\mathrm{tied}}$ indicates ratios of means and variances are tied $\lambda_z^\mu=\lambda_z^\sigma$.
We observe that sharing the ratios in SN degenerates performance.

\textbf{Subsets of $\boldsymbol{\Omega}$.}
Now we repeat training ResNet50 with SN$(8,32)$ three times but removing one normalizer in $\Omega$ at a time, resulting in three subsets \{IN,LN\}, \{IN,BN\}, and \{LN,BN\}.
Their top1/top5 accuracies are $75.30/92.46$, $76.12/92.86$ and $76.40/92.92$ respectively. They are comparable or better than just using IN, LN, and BN uniformly as reported in Table \ref{tab:comp}.
However, the result is $77.1/93.3$ when $\Omega=\{\mathrm{IN,LN,BN}\}$.
We conclude that
all normalizers in $\Omega$ are important to achieve the best accuracy.

\begin{table}[t]
\RawFloats
\centering
\scriptsize
\begin{tabular}%{l|cccccc}
{p{15pt}<{\centering}|p{32pt}<{\centering}p{28pt}<{\centering}p{28pt}<{\centering}p{28pt}<{\centering}
p{4pt}<{\centering}p{28pt}<{\centering}}
  \hline
  %&
 Res50 &BN \cite{C:BN}&GN \cite{GN}&IN \cite{IN}&LN \cite{LN}&SN&SN$_{\mathrm{tied}}$\\
  \hline
 $(8,32)$ & 76.4(-0.7) &75.9(-1.2)& 71.6(-5.5)& 74.7(-2.4)& \textbf{77.1} & 76.2(-0.9)\\
 $(8,8)$ &75.2(-1.5)& 76.0(-0.7) & --&-- & \textbf{76.7} &  76.1(-0.6) \\
 $(8,2)$ &65.3(-10.3) & \textbf{75.9}(0.3)& --& --& 75.6& 75.5(-0.1)\\
  \hline
 \end{tabular}
 \vspace{-5pt}
\caption{\small{Comparisons of top1 accuracy in ImageNet where ResNet50 is trained with different normalization methods. For example, $(8,32)$ indicates training with 8 GPUs and 32 samples per GPU, where means and variances are estimated in each GPU while gradients are aggregated across all GPUs. The best score is bold. The bracket reports difference between a method and SN. SN$_{\mathrm{tied}}$ indicates ratios of means and variances are tied.\vspace{-5pt}
}}\label{tab:comp}
\end{table}

\begin{table}[t]
\begin{subtable}{0.45\textwidth}
\centering
\scriptsize
\begin{tabular}%{l|cc}
{l|p{26pt}p{26pt}}
  \hline
 $(8,32)$ &ResNet50 &ResNet101 \\
  \hline
 BN & 76.4/93.0 & 77.6/93.6 \\%& 78.1/93.9 \\
 SN &77.1/93.3& 78.6/94.1 \\%& 78.3/94.0 \\
 SN$-$BN &0.7/0.3 & 1.0/0.5 \\
  \hline
 \end{tabular}
\end{subtable}
\hspace{10pt}
%\\
%\par\bigskip
\begin{subtable}{0.45\textwidth}
\centering
\scriptsize
\begin{tabular}%{l|cc}
{c|p{22pt}p{24pt}}
\hline
epoch& scratch & finetune\\
\hline
30$^{\mathrm{th}}$ & 76.5/93.0 & 77.4/93.4 \\
60$^{\mathrm{th}}$ & 76.1/93.0 & 77.1/93.4 \\
90$^{\mathrm{th}}$ & 76.2/93.0 & 76.9/93.3\\
\hline
\end{tabular}
\end{subtable}
\vspace{-5pt}
\caption{\small{\textbf{Left:} comparisons of top1/top5 accuracy of BN and SN trained with $(8,32)$ in ImageNet. ``SN-BN'' is the difference between their results. \textbf{Right:} comparisons of top1/top5 accuracy between `training from scratch' and `finetuning' ResNet50+SN$(8,32)$ with hard ratios until 100 epochs.\vspace{-5pt}
}}\label{tab:comp2}
\end{table}

\begin{figure}
  \centering
%\hspace{-22pt}
\includegraphics[width=1.1\linewidth]{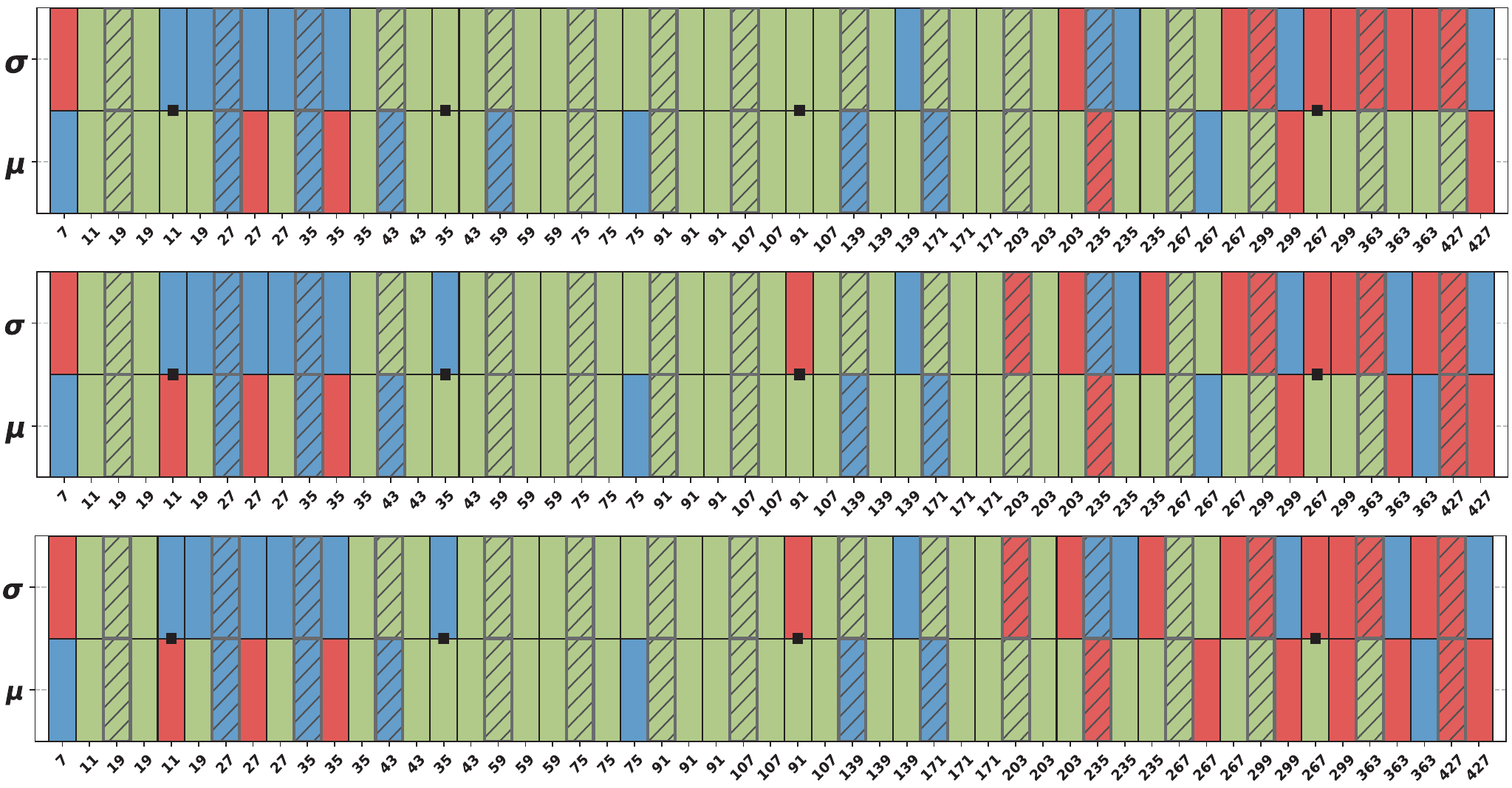}\\
\vspace{-10pt}
\caption{{\small \textbf{Hard ratios} for variance ($\sigma$) and mean ($\mu$) including BN ({\color{green}\textbf{green}}), IN ({\color{blue}\textbf{blue}}), and LN ({\color{red}\textbf{red}}). Snapshots of ResNet50 trained after 30 (\textbf{top}), 60 (\textbf{middle}), and 90 (\textbf{bottom}) epochs are shown.
RF is given for each layer (53 normalization layers in total). A bar with slashes denotes SN after $3\times3$ conv layer (the others are $1\times1$ conv). A black square `$\blacksquare$' indicates SN at the shortcut. It's better to zoom in 200\%.\vspace{-5pt}
}}\label{fig:SN8-32-epoch}
\end{figure}

\begin{figure*}
  \centering
  \includegraphics[width=1.0\linewidth]{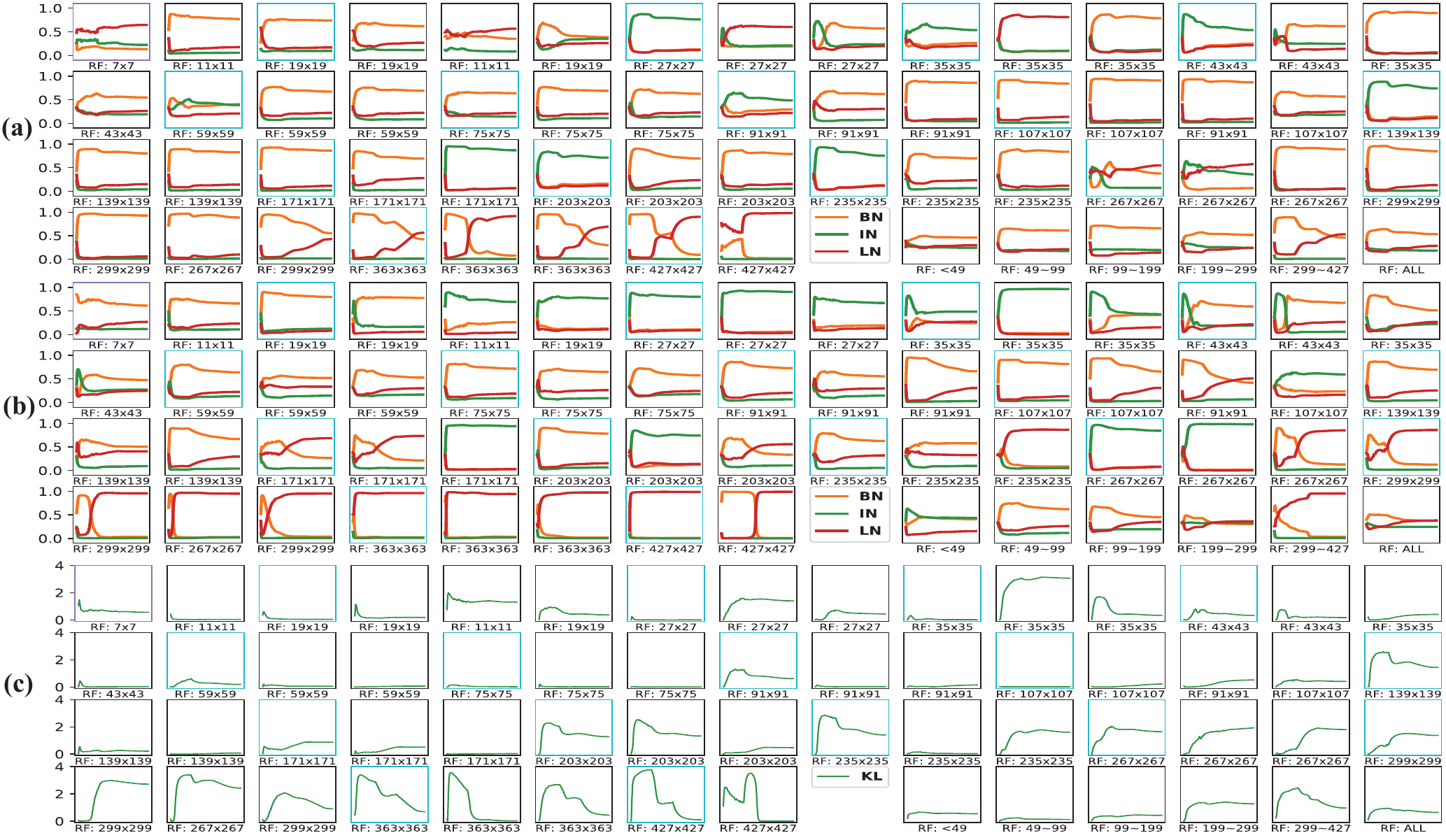}\\
  \caption{\small{Ratios of (a) $\lambda_z^\mu$ and (b) $\lambda_z^\sigma$ in {ResNet50+SN(8,32)} for each normalization layer for 100 epochs, as well as (c) their divergence $\mathcal{D}(\lambda_z^\mu\|\lambda_z^\sigma)$.
  Receptive field (RF) of each layer is given (53 normalization layers in total).
  The last 6 subfigures at the 4$^{\mathrm{th}}$, 8$^{\mathrm{th}}$, and 12$^{\mathrm{th}}$ row show results of different ranges of RF including `RF$<$49', `49$\sim$99', `99$\sim$199', `199$\sim$299', `299$\sim$427', and `ALL' (\ie 7$\sim$427).
 }}\label{fig:SN8-32-mu-sig}
\end{figure*}

\textbf{Random initializations.}
Here we show that the improvements of performance come from distinct normalizers but not random parameter initializations.
This is demonstrated by training ResNet50+SN(8,32) three times. At each time, we use a different random seed to initialize the network parameters.
The top1/top5 accuracies are $76.9/93.3$, $77.1/93.3$, and $76.9/93.4$ respectively.
Their average is $76.943/93.333\pm0.131/0.081$ that only has a small variance ($0.1\%$), demonstrating that results in Table \ref{tab:comp} are significant.

\subsubsection{Learning Dynamics of Ratios}

This section studies learning behaviors of ratios.

\textbf{Soft ratios vary in training.}
The values of soft ratios $\lambda_z^\mu$ and $\lambda_z^\sigma$ are between $0$ and $1$.
Fig.\ref{fig:SN8-32-mu-sig}(a,b) plot their values for each normalization layer at every epoch.
These values would have smooth fluctuation in training, implying that different epochs may have their own preference of normalizers.
In general, we see that $\lambda_z^\mu$ mostly prefers BN, while $\lambda_z^\sigma$ prefers IN and BN when receptive field (RF) $<$49 and prefers LN when RF$>$299.
Fig.\ref{fig:SN8-32-mu-sig}(c) shows the discrepancy between (a) and (b) by computing a symmetry metric, that is, $\mathcal{D}(\lambda_z^\mu\|\lambda_z^\sigma)=\mathcal{KL}(\lambda_z^\mu\|\lambda_z^\sigma)
+\mathcal{KL}(\lambda_z^\sigma\|\lambda_z^\mu)$ where $\mathcal{KL}(\cdot\|\cdot)$ is Kullback-Leibler divergence.
Larger value of $\mathcal{D}(\lambda_z^\mu\|\lambda_z^\sigma)$ indicates larger discrepancy between the distributions of $\lambda_z^\mu$ and $\lambda_z^\sigma$.

For example, $\lambda_z^\mu$ and $\lambda_z^\sigma$ of the first layer choose different normalizers (see the first subfigure in (a,b)), making them had moderately large divergence $\mathcal{D}(\lambda_z^\mu\|\lambda_z^\sigma)\approx1$ (see the first subfigure in (c)).
% as shown in (c).
Moreover, the ratios of the 2$^{\mathrm{nd}}$, 3$^{\mathrm{rd}}$, and 4$^{\mathrm{th}}$ layer are similar and they have small divergence $\mathcal{D}(\lambda_z^\mu\|\lambda_z^\sigma)<0.5$.
We also see that $\lambda_z^\mu$ and $\lambda_z^\sigma$ prefer different normalizers when RF is 199$\sim$299 and 299$\sim$427 where $\mathcal{D}(\lambda_z^\mu\|\lambda_z^\sigma)>2$, as shown in
the last row of (c).
In conclusion, more than 50\% number of layers have large divergence between $\lambda_z^\mu$ and $\lambda_z^\sigma$, confirming that different ratios should be learned for $\mu$ and $\sigma$.

\textbf{Hard ratios are relatively stable,} although the soft ratios are varying in training.
A hard ratio is a sparse vector obtained by applying $\mathrm{max}$ function to $\lambda_z^\mu$ or $\lambda_z^\sigma$ such as $\mathrm{max}(\lambda_z^\sigma)$, that is, only one entry is `1' and the others are `0' to select only one normalizer.
Fig.\ref{fig:SN8-32-epoch} shows hard ratios for each layer in three snapshots, which are ResNet50+SN(8,32) trained after 30, 60, and 90 epochs respectively.
For example, $\sigma$ and $\mu$ use LN and IN respectively in the first layer in Fig.\ref{fig:SN8-32-epoch}.

We have several observations. First, the size of filters ($3\times3$ or $1\times1$) seem to have no preference of specific normalizer, while the skip-connections prefer different normalizers for $\sigma$ and $\mu$ at the 90$^{\mathrm{th}}$ epoch.
Second, around $50\%$ number of layers select two different normalizers for $\sigma$ and $\mu$ in these three snapshots, which are $49\%$ ($26/53$), $53\%$ ($28/53$), and $51\%$ ($27/53$) respectively.
Third, the discrepancy between snapshots are small. For instance, $10$ layers are different when comparing between 30$^{\mathrm{th}}$ and 60$^{\mathrm{th}}$ epoch, while only $2$ layers are different between 60$^{\mathrm{th}}$ and 90$^{\mathrm{th}}$ epoch.
Fourth, the layers that choose different normalizers are mainly presented when RF$<$40 and $>$200, rendering depth would be a major factor that affects the ratios.

\begin{figure}
  \centering
  \includegraphics[width=.95\linewidth,height=100mm]{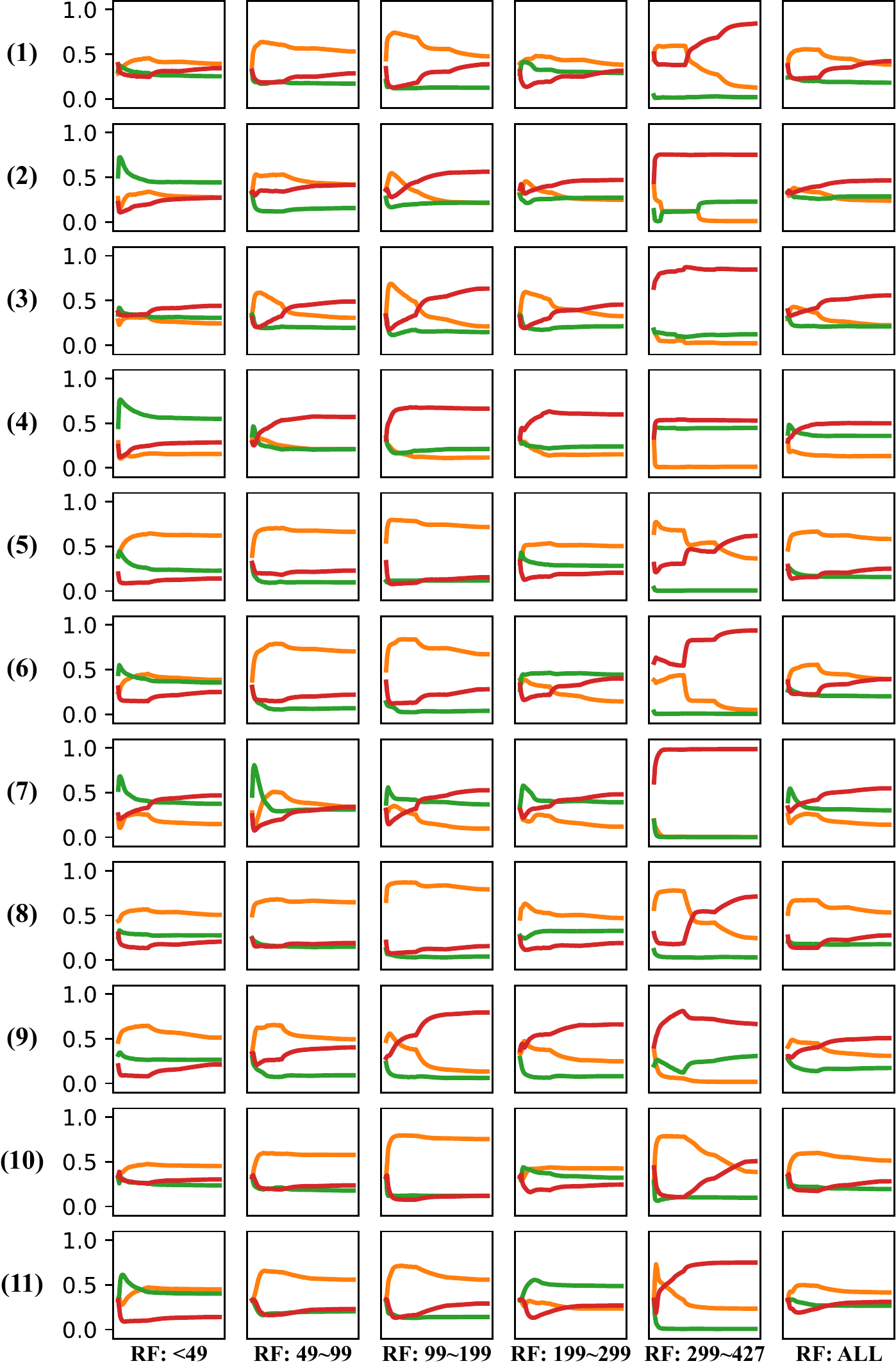}
  \caption{{\small \textbf{Comparisons of ratios} in ResNet50 trained with different settings in terms of ranges of receptive fields. BN, IN, and LN are visualized in {\color{orange}\textbf{orange}}, {\color{green}\textbf{green}}, and {\color{red}\textbf{red}}. \textbf{(1-2)} show $\lambda_z^\mu$ and $\lambda_z^\sigma$ of SN(8,8) respectively. \textbf{(3-4)} are $\lambda_z^\mu$ and $\lambda_z^\sigma$ of SN(8,2). \textbf{(5-7)} are ratios of $(8,32)$, $(8,8)$, and $(8,2)$ of SN$_{\mathrm{tied}}$ respectively ($\lambda_z^\mu=\lambda_z^\sigma$ in SN$_\mathrm{tied}$). \textbf{(8-9)} plot $\lambda_z^\mu$ and $\lambda_z^\sigma$ of downsampled-ImageNet. \textbf{(10-11)} plot $\lambda_z^\mu$ and $\lambda_z^\sigma$ of half-ImageNet.\vspace{-8pt}
  %\textbf{(12-17)} show $\lambda_z^\mu$ and $\lambda_z^\sigma$ of SN with subsets of normalizers \ie \{IN,LN\}, \{IN,BN\}, and \{LN,BN\}.
  }}\label{fig:RF}
\end{figure}

\textbf{Performance of hard ratios.}
We further examine performance of hard ratios.
We finetune the above snapshots by replacing soft ratios with hard ratios.
The right of Table \ref{tab:comp2} shows that these models perform as good as their soft counterpart ($77.1/93.3$), implying that we could increase sparsity in ratios to reduce computations of multiple normalizers while maintaining good performance.

Furthermore, we train the above models from scratch by initializing the ratios as the hard ratios in the above snapshots, rather than using the default initial value $\frac{1}{3}$.
However, this setting harms generalization as shown in Table \ref{tab:comp2}. We conjecture that all normalizers in $\Omega$ are helpful to smooth loss landscape at the beginning of training. Therefore, the sparsity of ratios should be enhanced gently as training progresses.
In other words, initializing ratios by $\frac{1}{3}$ is a good practice. Tuning this value is cumbersome and may also imped generalization ability.

\textbf{Depth is a main factor.}
%also
As shown by the ratios in Fig.\ref{fig:SN8-32-epoch}, IN mainly takes place in lower layers to reduce variations in low-level features, LN presents in deeper layers to increase learning ability, while BN presents in the middle.
Similar trait can be observed in Appendix Fig.\ref{appfig:SN8-8-2-con} showing ratios at the 100$^{\mathrm{th}}$ epoch when training converged.
Furthermore, we find that analogous trends can be also observed in the other networks such as ResNet101 and Inceptionv3 shown in Appendix Fig.\ref{appfig:in-101}.

\begin{figure}
  \centering
  \includegraphics[width=0.95\linewidth,height=80mm]{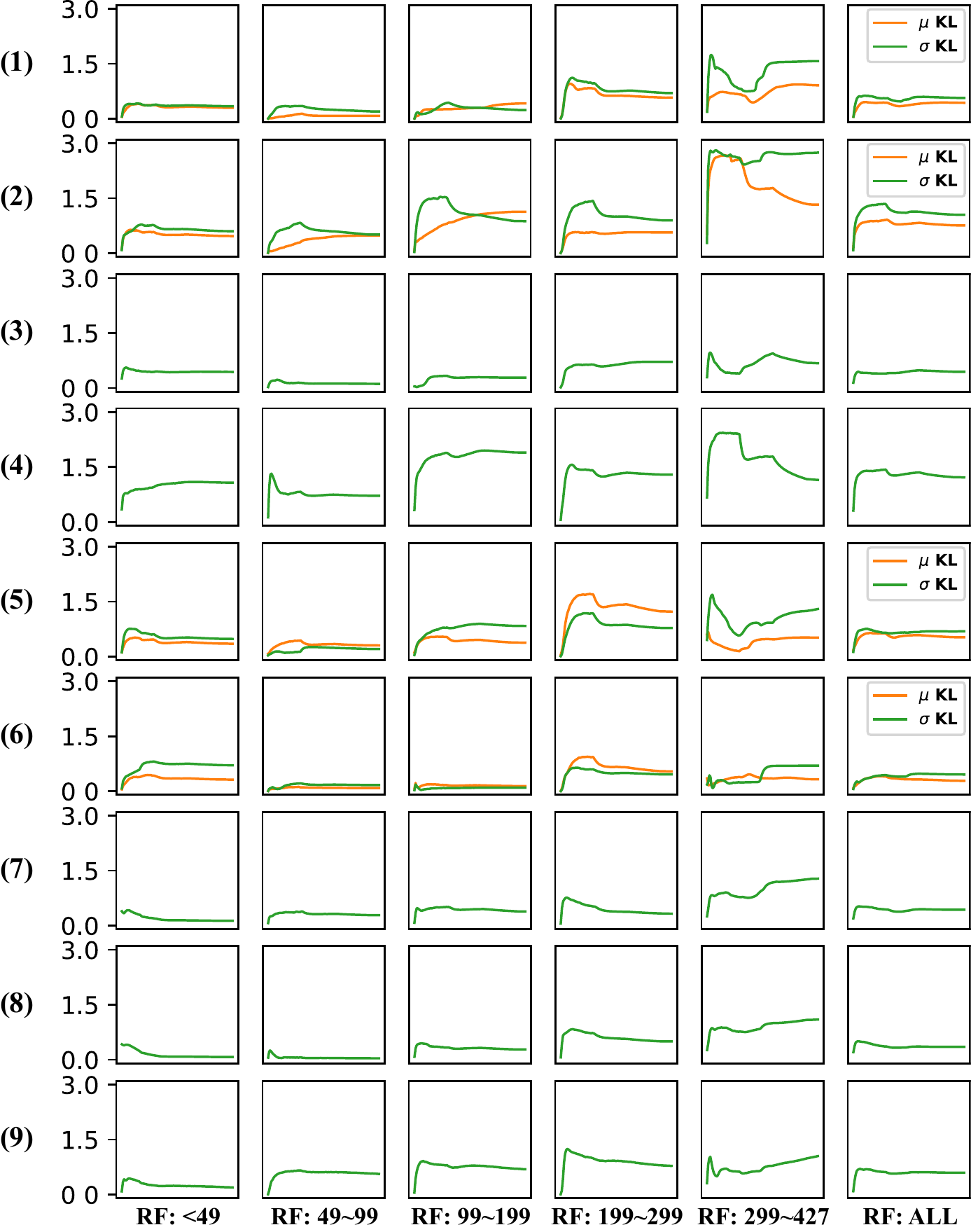}\\
  \caption{{\small \textbf{Divergence} between the ratios of ResNet50 trained with two different settings. \textbf{(1)} shows $\mathcal{D}({\lambda_z^\mu}_{\mathrm{SN}(8,32)}\|{\lambda_z^\mu}_{\mathrm{SN}(8,8)})$ and $\mathcal{D}({\lambda_z^\sigma}_{\mathrm{SN}(8,32)}\|{\lambda_z^\sigma}_{\mathrm{SN}(8,8)})$. Similarly, the others are \textbf{(2)} SN(8,32) \vs SN(8,2), \textbf{(3)} SN$_\mathrm{tied}$(8,32) \vs SN$_\mathrm{tied}$(8,8), \textbf{(4)} SN$_\mathrm{tied}$(8,32) \vs SN$_\mathrm{tied}$(8,2). Note that $\lambda_z^\mu=\lambda_z^\sigma$ in SN$_\mathrm{tied}$. \textbf{(5)} full- \vs downsampled-ImageNet. \textbf{(6)} full- \vs half-ImageNet. \textbf{(7)} RMSProp \vs SGD. \textbf{(8)} random initializations. \textbf{(9)} stepwise \vs cosine decay.\vspace{-8pt}
  }}\label{fig:RF-KL}
\end{figure}

\textbf{Batch size is another major factor.} We compare ratios by decreasing the batch size from $(8,32)$ to $(8,2)$. Results are shown in Fig.\ref{fig:RF}(1-7).
For models trained by SN, the ratios of BN are reduced at every RF range, see (1-4) and Fig.\ref{fig:SN8-32-mu-sig}(a,b).
Similar trend can be observed in (5-7) for SN$_\mathrm{tied}$.
These plots show that the dynamics of ratios are closely related to batch size, where BN would be reduced as it is unstable in small batch.
This can be also viewed by comparing Fig.\ref{fig:RF-KL}(1-4), where smaller batch size leads to larger divergence for both SN and SN$_\mathrm{tied}$.

\textbf{Input size and sample size are subordinate factors.} Here we investigate whether reducing image size and number of training data changes the dynamics of ratios.
Fig.\ref{fig:RF}(8-11) show the ratios of these two variants. In particular, ResNet50+SN(8,32) is trained by down-sampling each image from $224\times224$ to $64\times64$ (8-9), and trained using only 50\% ImageNet data while keeping the previous input size (10-11).

The top1/top5 accuracies of the above two variants are $55.90/78.78$ and $72.24/90.64$ respectively. It is seen that reducing image size degenerates accuracy in ImageNet more severely than reducing sample size by half.
To closely see this, we compare divergences of their ratios to the previous model that achieves $77.1/93.3$. As shown in Fig.\ref{fig:RF-KL}(5-6),
divergences in (5) are generally larger than (6) in every RF range.

\textbf{Solvers, random initializations, and learning rates are less relevant factors.}
Fig.\ref{fig:RF-KL}(7-9) examine the ratios in three factors including  solver (SGD \vs RMSProp), random parameter initialization, and learning rate decay (stepwise \vs cosine decay).
For the first two factors in (7-8), we find that their divergences in `RF:ALL' are mostly smaller than 0.5, while
the divergences in (9) are moderately larger than 0.5 showing that solver may marginally affect the ratios.

\begin{figure}[t]
%\vspace{-10pt}
%\hspace{-25pt}
\centering
	\begin{subfigure}{1\textwidth}
%\centering
\includegraphics[width=\textwidth,height=8mm]{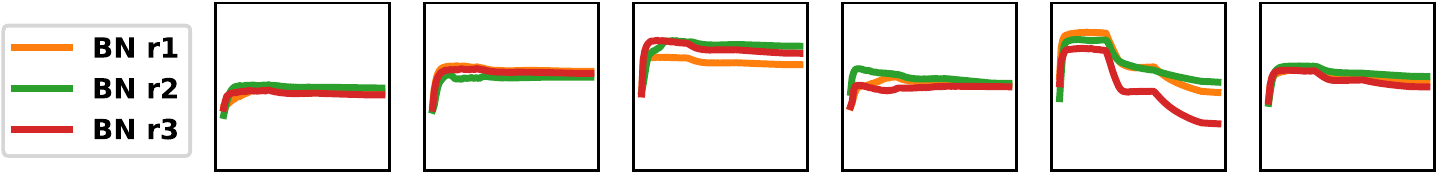}
		\vspace{-12pt}
	\end{subfigure}

\hspace{32pt}
	\begin{subfigure}{0.85\textwidth}
%\centering
\includegraphics[width=\textwidth,height=9mm]{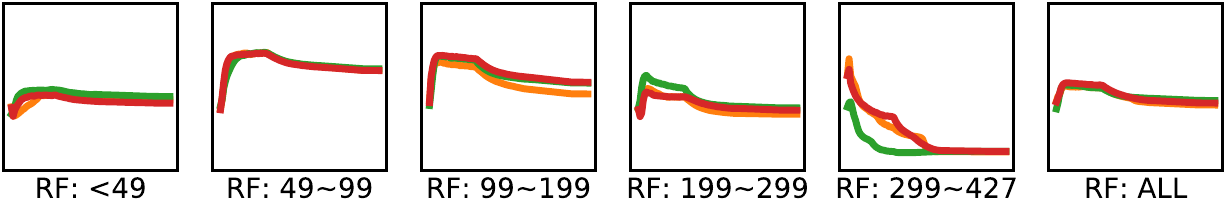}
		\vspace{-10pt}
	\end{subfigure}

\centering
	\begin{subfigure}{1\textwidth}
%\centering
\includegraphics[width=\textwidth,height=8mm]{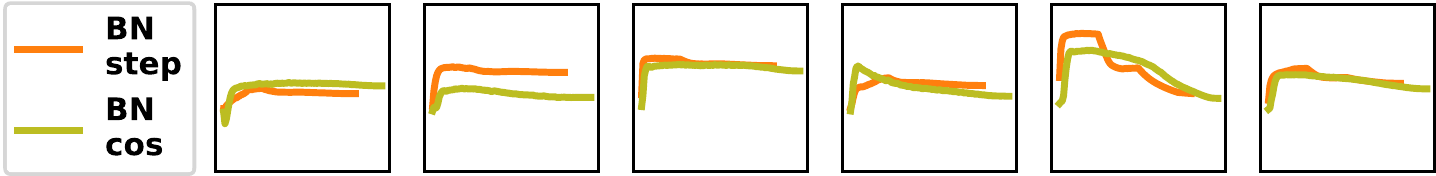}
		\vspace{-12pt}
	\end{subfigure}

\hspace{32pt}
	\begin{subfigure}{0.85\textwidth}
%\centering
\includegraphics[width=\textwidth,height=9mm]{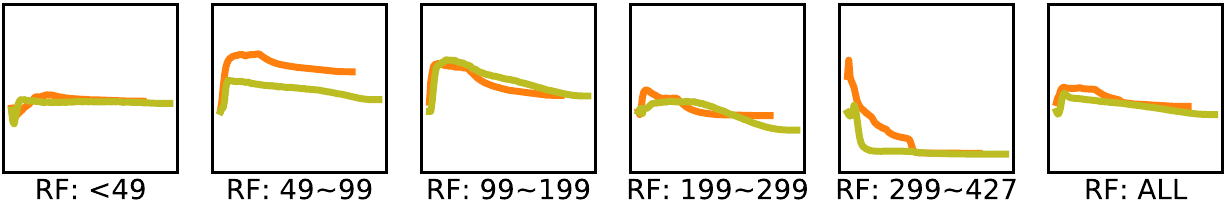}
		\vspace{-10pt}
	\end{subfigure}

	\begin{subfigure}{1\textwidth}
%\centering
\includegraphics[width=\textwidth,height=8mm]{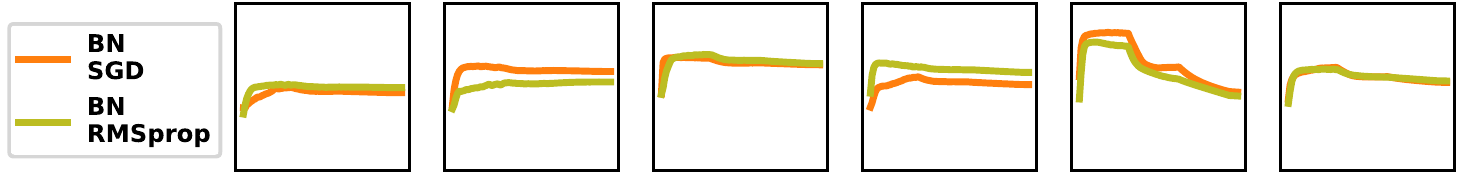}
		\vspace{-12pt}
	\end{subfigure}

\hspace{35pt}
\begin{subfigure}{0.84\textwidth}
%\centering
\includegraphics[width=\textwidth,height=9mm]{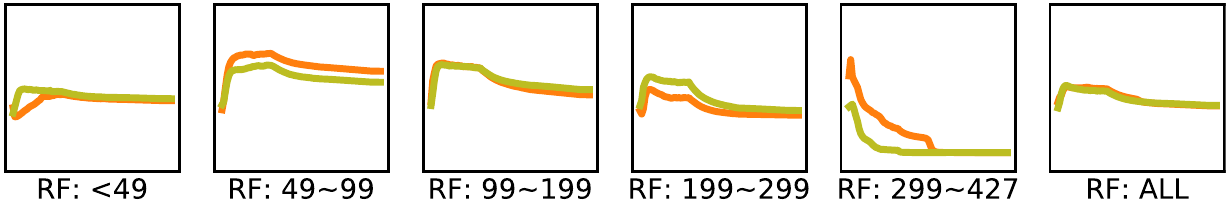}
		\vspace{-10pt}
	\end{subfigure}
\vspace{-5pt}
\caption{\small{\textbf{Top:} {ratios of BN} trained by randomly initializing parameters three times denoted as `random1', `random2', and `random3' for ResNet50+SN(8,32).
\textbf{Middle:} stepwise decay \vs cosine decay. \textbf{Bottom:} SGD \vs RMSProp. \vspace{-8pt}
}}
\label{fig:SN_lr_solver}
\end{figure}

More specifically, these three factors have more impact in upper layers such as RF$>$299 than lower layers. But they are not the main factors that affect the dynamics.
For example, the BN ratios are compared in Fig.\ref{fig:SN_lr_solver} where the difference among three random initializations are small, while the ratios are also similar for different solvers and learning rate decays.

\subsection{Object Detection and Scene Segmentation}

Now we use distinct normalizers in detection and segmentation.
To obtain representative results, we employ advanced frameworks such as Mask R-CNN \cite{MaskRCNN} for detection in COCO \cite{lin2014microsoft} and DeepLabv2 \cite{J:DeepLab} for segmentation in Cityscapes \cite{C:cityscape} and ADE20K \cite{A:ADE20K}. All these frameworks use ResNet50 as backbone and they are trained by following protocols in the corresponding benchmarks, while only the normalization layers are replaced by SN.
More empirical settings are provided in Appendix \ref{app:exp_proto}.

\textbf{Comparisons.}
Table \ref{tab:det_seg} compares different models that are pretrained by using BN \cite{C:BN}, GN \cite{GN}, and SN with three batch sizes.
All these models are finetuned by $(8,2)$, which is a popular setup in these benchmarks. Moreover, by following usual practice, BN is frozen in COCO and it is synchronized across 8 GPUs in Cityscapes and ADE20K, while GN and SN are neither frozen nor synchronized.

We see that using BN and GN uniformly in the network does not perform well in these tasks, even though BN is frozen or synchronized.
In COCO and ADE20K, `SN$(8,2)$' performs significantly better than the others.
In these two tasks, `SN$(8,32)$' and `SN$(8,4)$' may reduce performance, because BN has large ratio in these models, implying that finetuning SN pretrained with large batch to small batch would be unstable.
Nevertheless, `SN$(8,32)$' and `SN$(8,4)$' achieve better results than `SN$(8,2)$' in Cityscapes.
This could be attributed to large input image size 713$\times$713 that diminishes noise in the batch statistics of BN (in SN).

\begin{figure}
  \centering
  \includegraphics[width=.9\linewidth,height=65mm]{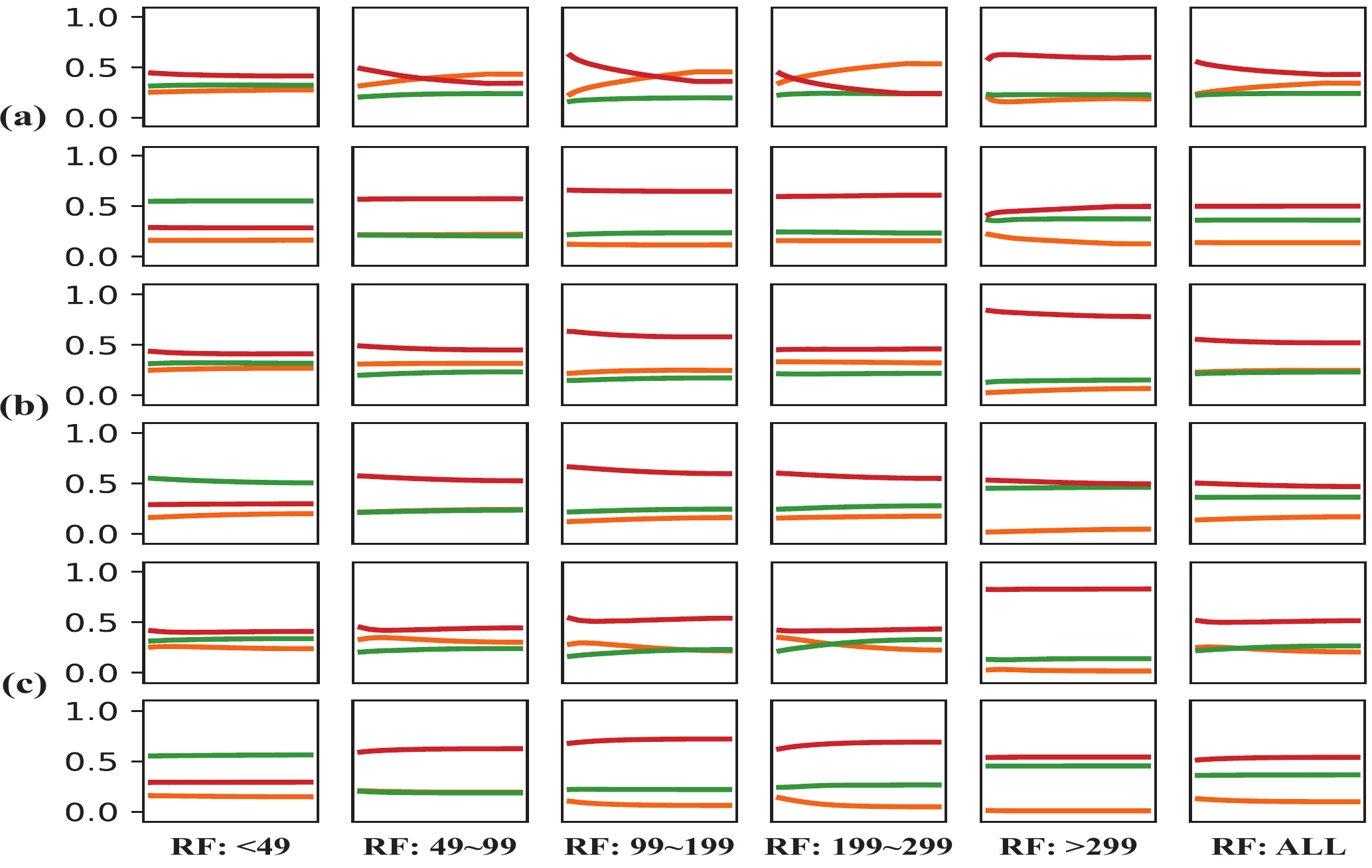}\\
  \caption{{\small \textbf{Ratios for detection and segmentation} including BN ({\color{orange}\textbf{orange}}), IN ({\color{green}\textbf{green}}), and LN ({\color{red}\textbf{red}}). We show $\lambda_z^\mu$ and $\lambda_z^\sigma$ in ResNet50+SN(8,2) finetuned to (a) COCO, (b) Cityscapes, and (c) ADE20K. \vspace{-8pt}
  }}\label{fig:SN_det_seg}
\end{figure}

\begin{table}[t]
\RawFloats
\centering
\scriptsize
\begin{tabular}%{l|cccccc}
{p{45pt}|p{26pt}p{26pt}p{26pt}p{26pt}p{18pt}}
  \hline
  %&
  & BN(8,32) & GN(8,32) & SN(8,32)& SN(8,4) & SN(8,2)\\
  \hline
 mAP$^{\mathrm{box}}_{\mathrm{COCO}}$ & 38.6/-2.4 &40.2/-0.8 &39.8/-1.2&--&\textbf{41.0}\\
 mAP$^{\mathrm{segm}}_{\mathrm{COCO}}$ & 34.2/-2.3 & 35.7/-0.8 & 35.3/-1.2 &--& \textbf{36.5}\\
 mIoU$_{\mathrm{Cityscapes}}$ &72.7/-3.1  & 72.2/-3.6 & \textbf{75.8}/+0.4 & \textbf{75.8}/+0.4 & 75.4 \\
 mIoU$_{\mathrm{ADE20K}}$ &37.7/-1.5 & 36.3/-2.9 &38.4/-0.8& 39.0/-0.2 & \textbf{39.2}\\
  \hline
 \end{tabular}
\caption{\small{\textbf{Comparisons of detection and segmentation}. ResNet50 is backbone and pretrained with BN, GN, and different batch sizes of SN. We adopt Mask R-CNN for detection and DeepLabv2 for segmentation. All methods are finetuned in $(8,2)$ and evaluated in COCO 2017 val set, ADE20K val set, and Cityscapes test set respectively.
%The best result for each task is bold.
``$\cdot/\cdot$'' shows a result and its gap compared to SN$(8,2)$. \vspace{-8pt}
}}\label{tab:det_seg}
\end{table}

\begin{figure*}[t]
%\vspace{-10pt}
%\hspace{-25pt}
\centering
\includegraphics[width=\textwidth,height=100mm]{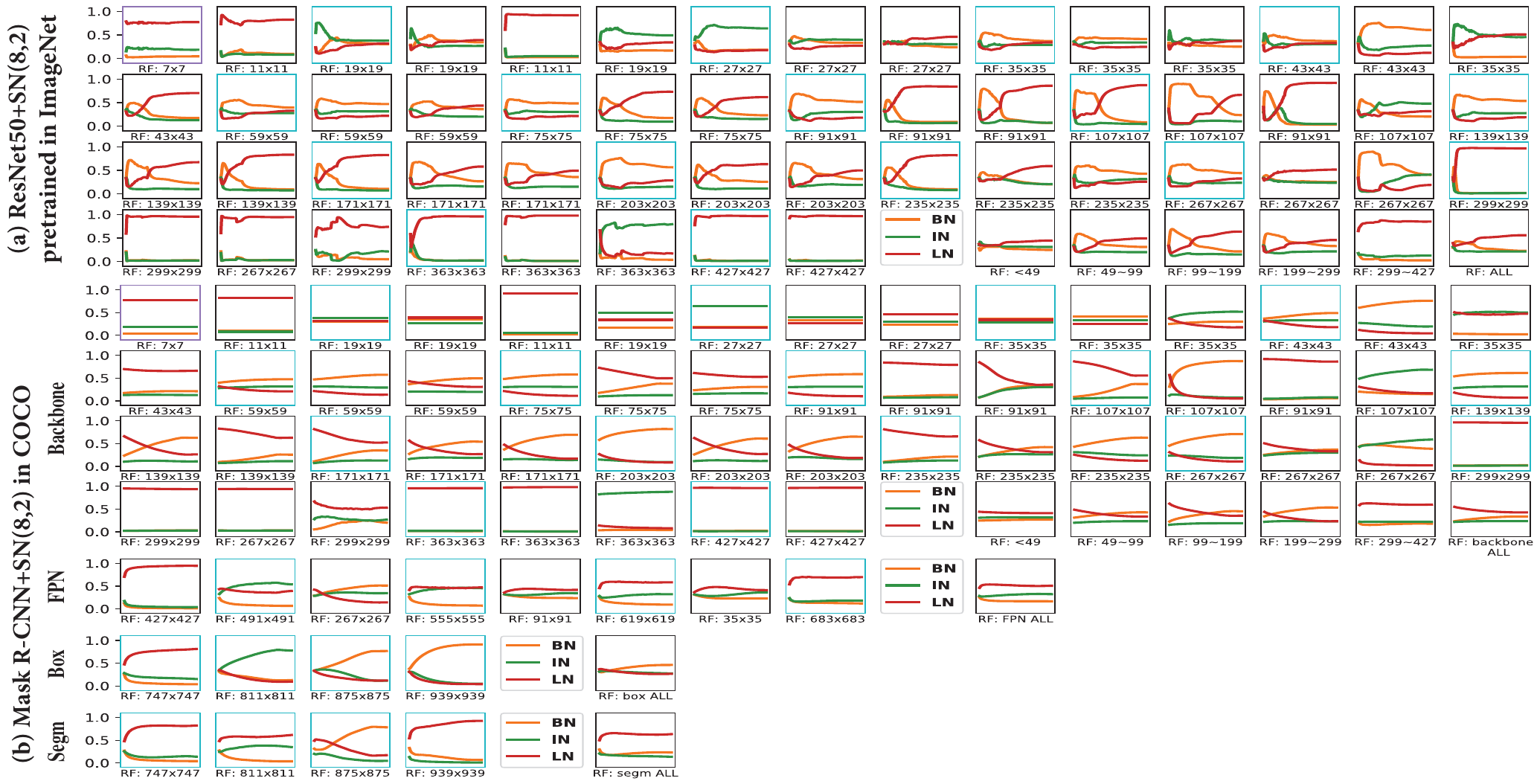}
\vspace{-15pt}
\caption{\small{(a) shows ratios of ResNet50+SN$(8,2)$ in ImageNet. (b) shows ratios of Mask R-CNN+SN$(8,2)$ when finetuning in COCO including backbone (ResNet50), FPN, box stream, and mask stream. \vspace{-8pt}
}}
\label{fig:det_coco}
\end{figure*}

\textbf{Dynamics of ratios are more smooth in finetuning than pretraining.}
Fig.\ref{fig:SN_det_seg} visualizes ratios in finetuning, which are more smooth than pretraining as compared to Fig.\ref{fig:RF}. Intuitively, this is because a small learning rate is typically used in finetuning.
In Fig.\ref{fig:SN_det_seg}, IN and LN ratios are generally larger than BN because of small batch size. The lower layers (RF$<$49) prefer IN more than the upper layers (RF$>$299) that choose LN.
When comparing different tasks, Fig.\ref{fig:SN_det_seg}(b,c) in segmentation have analogue dynamics where BN ratios are gently decreased ($<$0.3). But they are different from detection in (a) where BN gradually increases when 49$<$RF$<$299 ($>$0.5). This could be attributed to the two-stage pipeline of Mask R-CNN.
To see this, Fig.\ref{fig:det_coco}(b) and Appendix Fig.\ref{fig:det_converge} plot the ratios in COCO.
We see that BN has larger impact in backbone and box stream than the other components.
Furthermore, ratios in the backbone have different dynamics in pretraining and finetuning by comparing two ResNet50 models in Fig.\ref{fig:det_coco}(a,b), that is, the BN ratios decrease in pretraining for recognition, while increase in finetuning for detection even though the batch size is $(8,2)$.

\textbf{Finetuning appropriate pretrained models is a good practice.}
We observe that ratios pretrained with different batch sizes bring different impacts in finetuning.
Using models that are pretrained and finetuned with comparable batch size would be a best practice for good results.
Otherwise, performance may degenerate.

We take ADE20K as an example. Appendix Fig.\ref{appfig:seg_training_ade} shows the ratios when SN$(8,32)$ is used in pretraining but SN$(8,2)$ in finetuning. In line with expectation, the BN ratios are suppressed during finetuning, as the batch statistics become unstable.
However, these ratios are still suboptimal until training converged as shown in Appendix Fig.\ref{appfig:segm_converage_ade}, where the BN ratios finetuned from SN$(8,32)$ are still larger than those directly finetuned from SN$(8,2)$, reducing performance in ADE20K (see Table \ref{tab:det_seg}).

%\vspace{-5pt}
\section{Summary and Future Work}\label{sec:res}

This work studies a new viewpoint in deep learning, \emph{showing that each convolutional layer would be better to select its own normalizer}. We investigate ratios of normalizers in popular benchmarks including ImageNet, COCO, Cityscapes and ADE20K, and
summarize our findings.

\vspace{-8pt}
\begin{itemize}
\item Learning dynamics of ratios in ImageNet are more relevant to depth of networks, batch size, and image size, but less pertinent to random parameter initialization, learning rate decay, and solver.
\vspace{-6pt}
\item The ratios of BN are proportional to batch size, that is, BN ratios increase along with the increase of batch size. IN and LN are inversely proportional to batch size. LN is proportional to depth. BN and IN are inversely proportional to depth.
%, and BN ratios are proportional to batch size and inversely proportional to depth.
\vspace{-6pt}
\item Removing any one normalizer from $\Omega$ harms generalization in ImageNet. Similar trait has been observed in the other models such as ResNet101 and Inceptionv3.
\vspace{-6pt}
\item Hard (sparse) ratios could outperform soft ratios in SN. But the soft ratios may help smooth loss landscape, initializing the ratios by using hard ratios instead of $\frac{1}{3}$ harms generalization. In other words, sparsity of ratios should be enhanced during training, rather than at the very beginning of training.
\vspace{-6pt}
\item Recognition, detection, and segmentation have distinct learning dynamics of ratios.
%,
More important practice of SN, trial and error are summarized in Appendix \ref{appsec:error}.
\end{itemize}

Future work involve three aspects. (1) Algorithm will be devised to learn sparse ratios for SN.
(2) We will impose structure in ratios such as dividing them into groups to choose normalizer for each group.
(3) As IN and LN also work well in the tasks of low-level vision and recurrent neural networks (RNNs), trying SN in these problems is also a future direction.
(4) We'll try to understand the learning and generalization ability of SN theoretically, though it is an open and challenging problem in deep learning.
(5) Switching between whitening \cite{GWNN,EigenNet} and standardization (\eg BN) will be also important and valuable.

{\small
\bibliographystyle{ieee}
\bibliography{egbib}
}

\newpage
\appendix
\newpage

\section*{Appendices}
\addcontentsline{toc}{section}{Appendices}
\renewcommand{\thesubsection}{\Alph{subsection}}

\subsection{Relation with Meta Learning}\label{appsec:ML}

We draw a connection between SN's learning problem and meta learning (ML) \cite{bilevel,hyperparameter,ENAS}, which can be also used to learn the control parameters in SN. In general, ML is defined as $\min_{\Theta}\mathcal{L}(\Theta,\Phi^\ast)+\min_{\Phi}\varphi\mathcal{L}(\Theta^\ast,\Phi)$ where $\varphi$ is a constant multiplier.
Unlike SN trained with a single stage, this loss function is minimized by performing two feed-forward stages iteratively until converged.
First, by fixing the current estimated control parameters $\Phi^\ast$, the network parameters are optimized by $\Theta^\ast=\min_\Theta\mathcal{L}(\Theta,\Phi^\ast)$.
Second, by fixing $\Theta^\ast$, the control parameters are found by $\Phi^\ast=\min_\Phi\varphi\mathcal{L}(\Theta^*,\Phi)$.

The above two stages are usually optimized by using two different sets of training data. %
For example, previous work \cite{DARTS,ENAS} used $\Phi$ to search network architectures from a set of modules with different numbers of parameters and computational complexities. They divided an entire training set into a training and a validation set without overlapping, where $\Phi$ is learned from the validation set while $\Theta$ is learned from the training set.
This is because $\Phi$ would choose the module with large complexity to overfit training data, if both $\Theta$ and $\Phi$ are optimized in the same dataset.

The above 2-stage training increases computations and runtime.
In contrast, $\Theta$ and $\Phi$ for SN can be generally optimized within a single stage in the same dataset, because $\Phi$ regularizes training by choosing different normalizers from $\Omega$ to prevent overfitting.
%, 

\subsection{Trial \& Error}\label{appsec:error}

In Table \ref{tab:error}, we report several important practices when learning to select normalizers. Many of these practices shed light on future work of sparse SN and synchronized SN.

\begin{table}[t]
\small
  \centering
  \caption{{\small Summary of practices that help training CNNs with SN.}}
  \begin{tabular}{p{3pt}p{210pt}}
  \hline
  1. & Initializing ratios of normalizers uniformly \eg $1/3$. Carefully tuning the initial ratios may harm generalization. \\
  2. & Adding dropout with a small ratio (\eg 0.1$\sim$0.2) after each SN layer provides minor improvement of generalization in ImageNet, but it reduces over-fitting.\\
  3. & Adding $0.5$ dropout in the last fully-connected layer helps generalization in ImageNet.\\
  4. & A model in pretraining and finetuning should have comparable batch size.\\
  5. & Do not put SN after global pooling when feature map size is 1$\times$1, because IN and LN are unstable after global pooling.\\
  6. & SN$_{\mathrm{tied}}$ performs comparably well with SN when batch size is small \eg $(8,2)$.\\
  7. & Sparse SN improves SN in ImageNet.\\
  8. & Sparse SN reduces computational runtime in inference compared to SN. 50\% number of layers in sparse SN select BN for both $\mu$ and $\sigma$, meaning that these BN layers can be turned into linear transformation to reduce runtime in inference.\\
  9. & Synchronizing BN in SN improves generalization.\\
  \hline
  \end{tabular}
  \label{tab:error}
\end{table}

\subsection{Experimental Protocols}\label{app:exp_proto}

\textbf{ImageNet.}
All models in ImageNet are trained on 1.2M images and evaluated on 50K validation images. They are trained by using SGD with different settings of batch sizes, which are denoted as a 2-tuple, (\emph{number of GPUs}, \emph{number of samples per GPU}).
For each setting, the gradients are aggregated over all GPUs, and the means and variances of the normalization methods are computed in each GPU.
The network parameters are initialized by following \cite{C:resnet}.
For all normalization methods, all $\gamma$'s
are initialized as 1 and all $\beta$'s as 0.
The parameters of SN ($\lambda_z^\mu$ and $\lambda_z^\sigma$) are initialized as 1/3.
We use a weight decay of $10^{-4}$ for all parameters including $\gamma$ and $\beta$.
All models are trained for 100 epoches with a initial learning rate of 0.1, which is deceased by 10$\times$ after 30, 60, and 90 epoches.
For different batch sizes, the initial learning rate is linearly scaled according to \cite{ImageNet1hour}.
During training, we employ data augmentation the same as \cite{C:resnet}.
The top-1 classification accuracy on the 224$\times$224 center crop is reported.

\textbf{COCO.} We train all models on 8 GPUs and 2 images per GPU. Each image is re-scaled to its shorter side of 800 pixels.
%The train-time proposals for FPN are 2000, while the test-time proposals are 1000.
%
In particular, the learning rate (LR) is initialized as 0.02 and is decreased by the LR schedule as 2$\mathrm{x}$ schedule.
We set weight decay to 0 for both $\gamma$ and $\beta$ following \cite{GN}.

All the above models
are trained in the \emph{2017 train} set of COCO by using SGD with a momentum of 0.9 and a weight decay of $10^{-4}$ on the network parameters, and tested in the \emph{2017 val} set.
We report the standard metrics of COCO, that is, average precisions at IoU=0.5:0.05:0.75 (AP).
%,

\begin{figure*}[t]
%\vspace{-10pt}
%\hspace{-25pt}
\centering
\begin{subfigure}{0.95\textwidth}
\centering
\includegraphics[width=\textwidth,height=24mm]{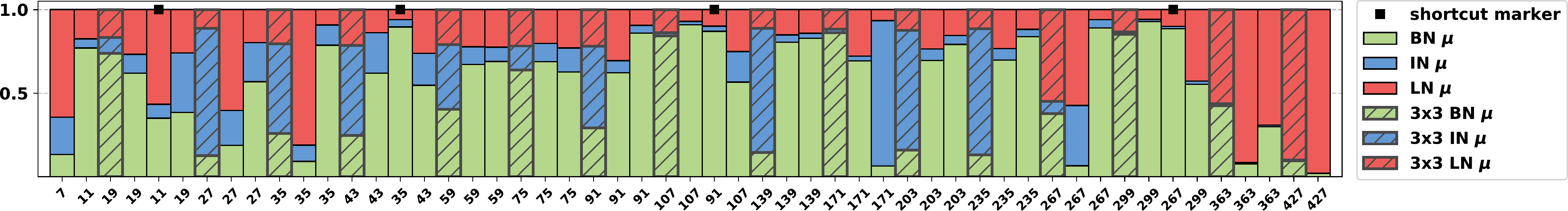}
		\vspace{-15pt}\caption{{\small $\lambda_z^\mu$ in \textbf{ResNet50+SN(8,32)} when pretraining converged in ImageNet.}
}
	\end{subfigure}
	~
	\begin{subfigure}{0.95\textwidth}
\centering
\includegraphics[width=\textwidth,height=24mm]{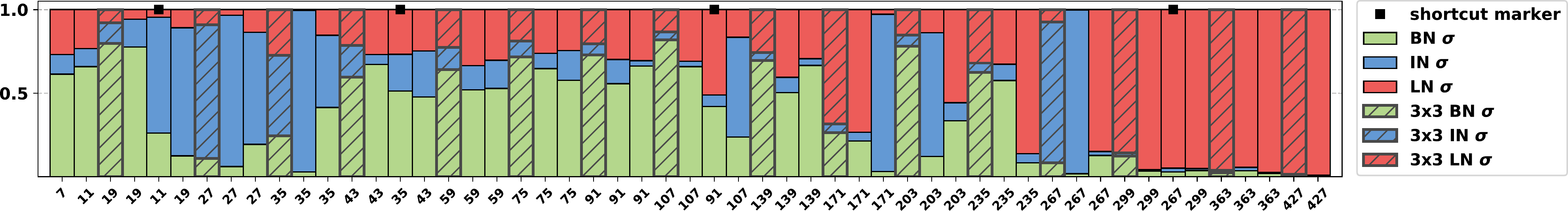}
		\vspace{-15pt}\caption{{\small $\lambda_z^\sigma$ in \textbf{ResNet50+SN(8,32)} when pretraining converged in ImageNet.}}
	\end{subfigure}
~
	\begin{subfigure}{0.95\textwidth}
\centering
\includegraphics[width=\textwidth,height=24mm]{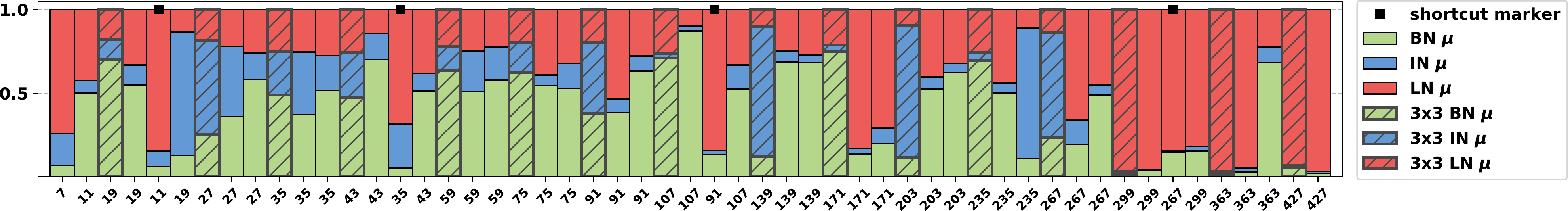}
		\vspace{-15pt}\caption{$\lambda_z^\mu$ in \textbf{ResNet50+SN(8,8)} when pretraining converged in ImageNet.
}
	\end{subfigure}
	~
	\begin{subfigure}{0.95\textwidth}
\centering
\includegraphics[width=\textwidth,height=24mm]{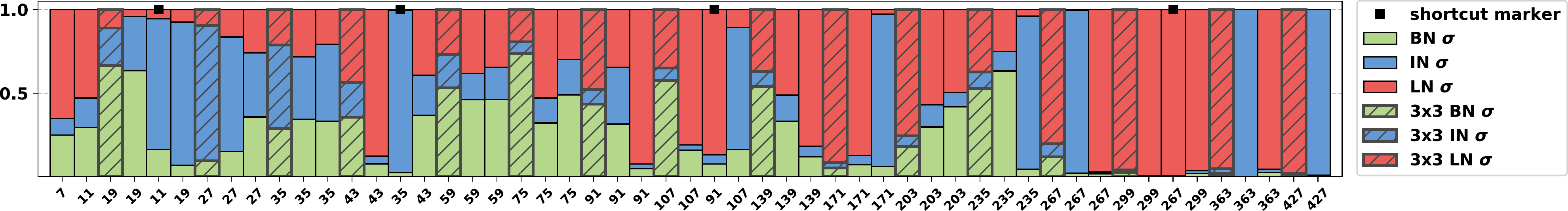}
		\vspace{-15pt}\caption{$\lambda_z^\sigma$ in \textbf{ResNet50+SN(8,8)} when pretraining converged in ImageNet. }
	\end{subfigure}
~
\begin{subfigure}{0.95\textwidth}
\centering
\includegraphics[width=\textwidth,height=24mm]{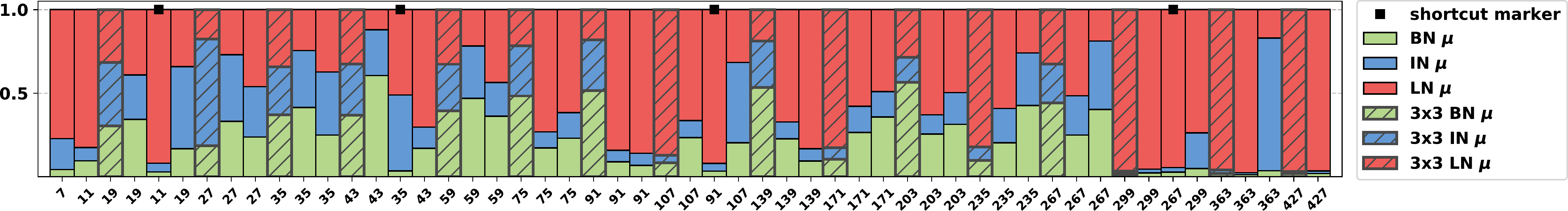}
		\vspace{-15pt}\caption{$\lambda_z^\mu$ in \textbf{ResNet50+SN(8,2)} when pretraining converged in ImageNet.}
	\end{subfigure}
~
\begin{subfigure}{0.95\textwidth}
\centering
\includegraphics[width=\textwidth,height=24mm]{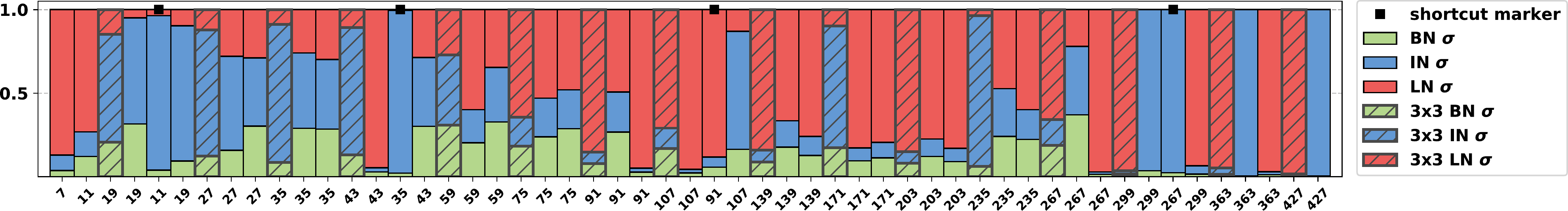}
		\vspace{-15pt}\caption{$\lambda_z^\sigma$ in \textbf{ResNet50+SN(8,2)} when pretraining converged in ImageNet.}
	\end{subfigure}
\vspace{-5pt}
\caption{{$\lambda_z^\mu$ and $\lambda_z^\sigma$ for ResNet50 in ImageNet when training converged. RF is given for each layer. A bar with slashes denotes SN after $3\times3$ conv layer (the others are $1\times1$ conv layer). A black square `$\blacksquare$' indicates SN at the shortcut.
}}
\label{appfig:SN8-8-2-con}
\end{figure*}

\textbf{Cityscapes and ADE20K.}
We use 2 samples per GPU for ADE20K and 1 sample per GPU for Cityscapes.
We employ the open-source software in PyTorch\footnote{\url{https://github.com/CSAILVision/semantic-segmentation-pytorch}} and only replace the normalization layers in CNNs with the other settings fixed. For both datasets, we use  DeepLabv2~\cite{J:DeepLab} with ResNet50 as the backbone network, where
%, using the original $7\times7$ kernel size in the first convolution layer.
%
$\mathrm{output}\_\mathrm{stride}=8$ and the last two blocks in the original ResNet contains atrous convolution with $\mathrm{rate}=2$ and $\mathrm{rate}=4$ respectively.
Following~\cite{C:PSPNet}, we employ ``poly" learning rate policy with $\mathrm{power}=0.9$ and use the auxiliary loss with the weight $0.4$ during training.
The bilinear operation is adopted to upsmaple the score maps in the validation phase.

In ADE20K,
we resize each image to $450\times450$ and train for $100,000$ iterations. We performance multi-scale testing with $\mathrm{input}\_\mathrm{size}=\{300,400,500,600\}$.
In Cityscapes,
we use random crop with the size $713\times713$ and train for $700$ epoches. For multi-scale testing, the inference scales are $\{1.0, 1.25, 1.5, 1.75\}$.

\subsection{More Results}

More results are plotted in the remaining figures due to the limited length of the paper.

\begin{figure*}[t]
%\vspace{-10pt}
%\hspace{-25pt}
\centering
	\begin{subfigure}{1\textwidth}
%\centering
\includegraphics[width=\textwidth,height=24mm]{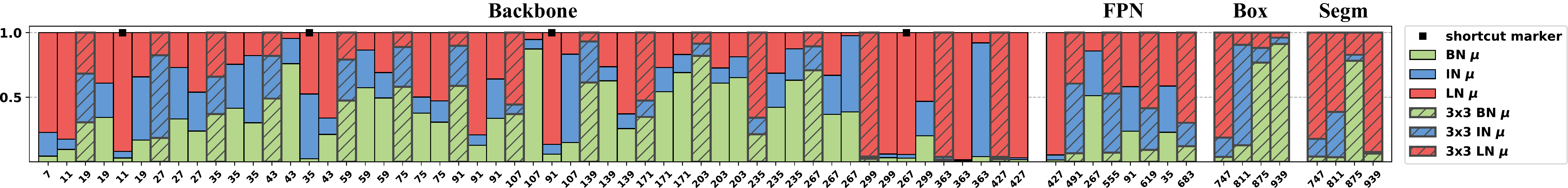}
		\vspace{-12pt}
	\end{subfigure}

%\hspace{32pt}
	\begin{subfigure}{1\textwidth}
%\centering
\includegraphics[width=\textwidth,height=24mm]{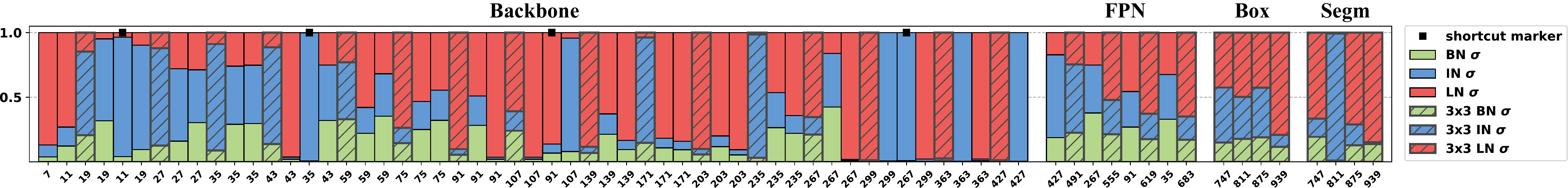}
		\vspace{-10pt}
	\end{subfigure}

\vspace{-5pt}
\caption{{$\lambda_z^\mu$ (top) and $\lambda_z^\sigma$ (bottom) of Mask R-CNN+SN(8,2) are shown when training converged in COCO, including backbone (ResNet50), FPN, box stream, and mask stream.
}}
\label{fig:det_converge}
\end{figure*}

\begin{figure*}
  \centering
  \includegraphics[width=.9\linewidth]{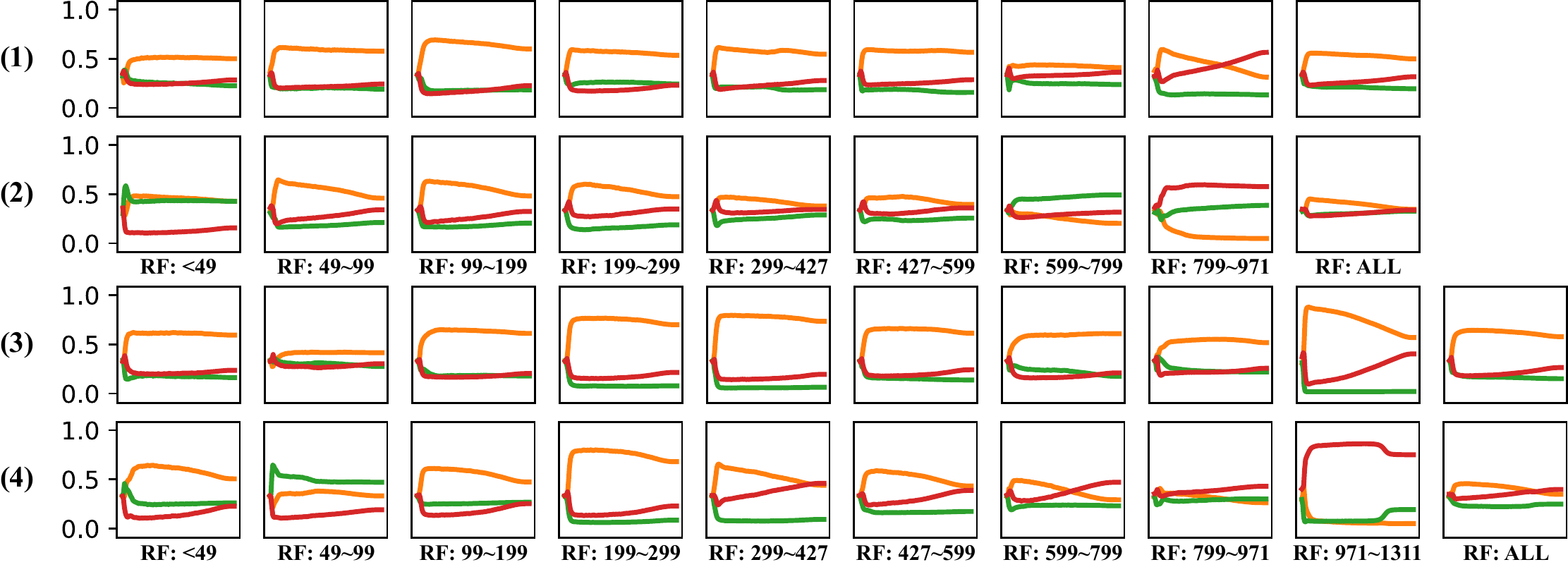}\\
  \caption{(1-2) plot $\lambda_z^\mu$ and $\lambda_z^\sigma$ for ResNet101+SN(8,32), while (3-4) show $\lambda_z^\mu$ and $\lambda_z^\sigma$ for Inceptionv3+SN(8,32).}
  \label{appfig:in-101}
\end{figure*}

\begin{figure*}[t]
%\vspace{-10pt}
%\hspace{-25pt}
\centering
	\begin{subfigure}{0.95\textwidth}
\centering
\includegraphics[width=\textwidth,height=30mm]{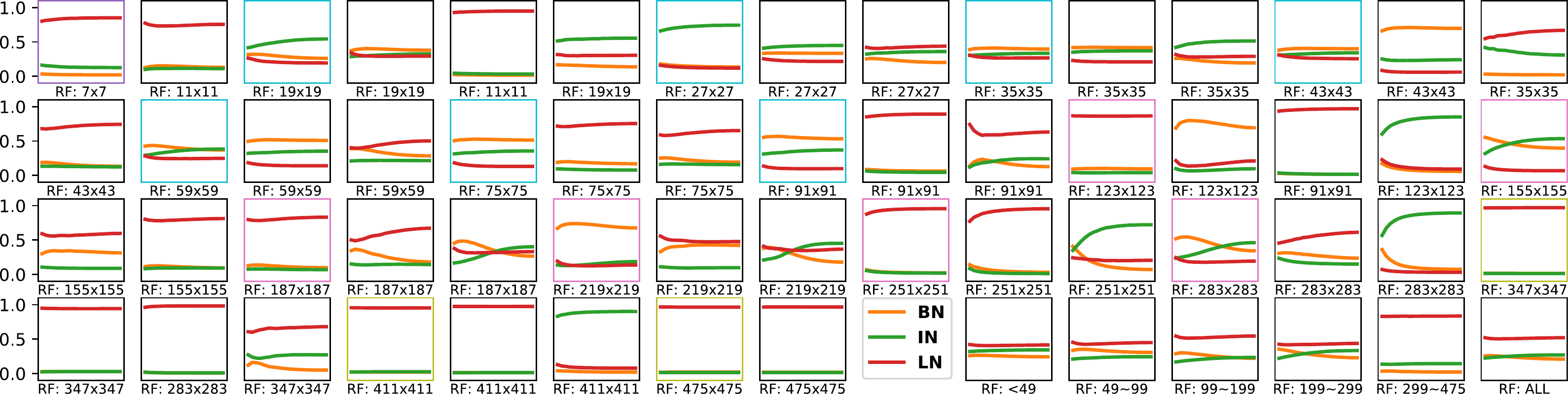}
		\vspace{-15pt}\caption{$\lambda_z^\mu$ in \textbf{SN(8,2)} for ADE20K.}
	\end{subfigure}
	~
	\begin{subfigure}{0.95\textwidth}
\centering
\includegraphics[width=\textwidth,height=30mm]{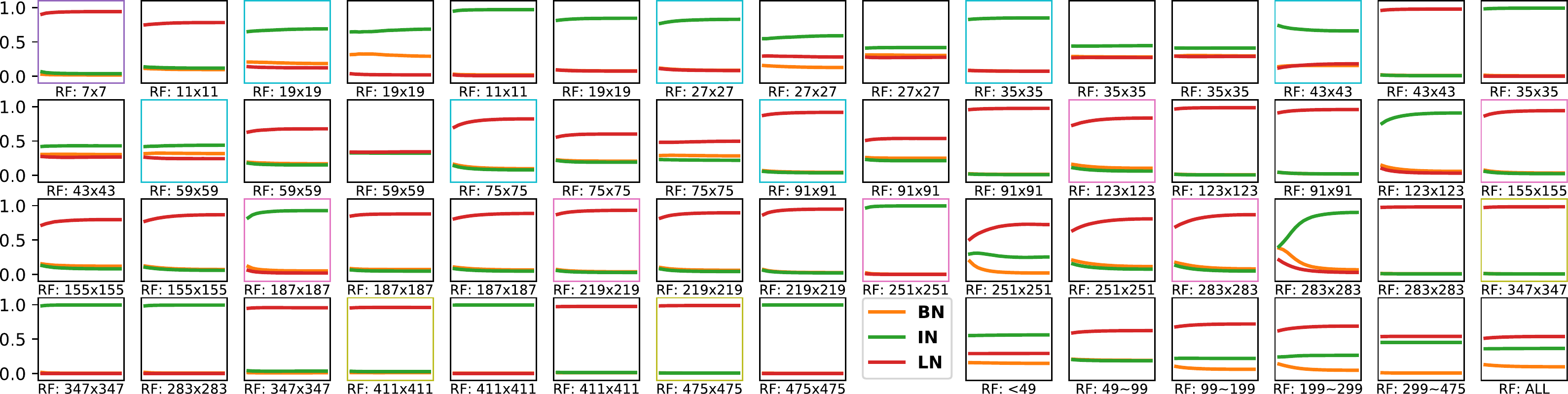}
		\vspace{-15pt}\caption{$\lambda_z^\sigma$ in \textbf{SN(8,2)} for ADE20K.}
	\end{subfigure}
~
\begin{subfigure}{0.95\textwidth}
\centering
\includegraphics[width=\textwidth,height=30mm]{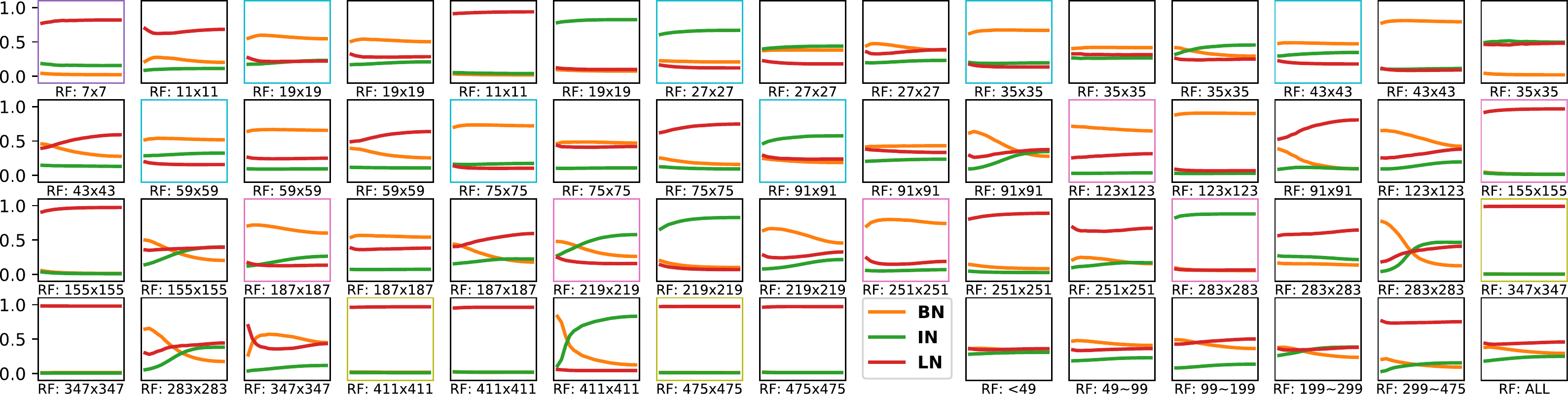}
		\vspace{-15pt}\caption{$\lambda_z^\mu$ in \textbf{SN(8,4)} for ADE20K.}
	\end{subfigure}
~
\begin{subfigure}{0.95\textwidth}
\centering
\includegraphics[width=\textwidth,height=30mm]{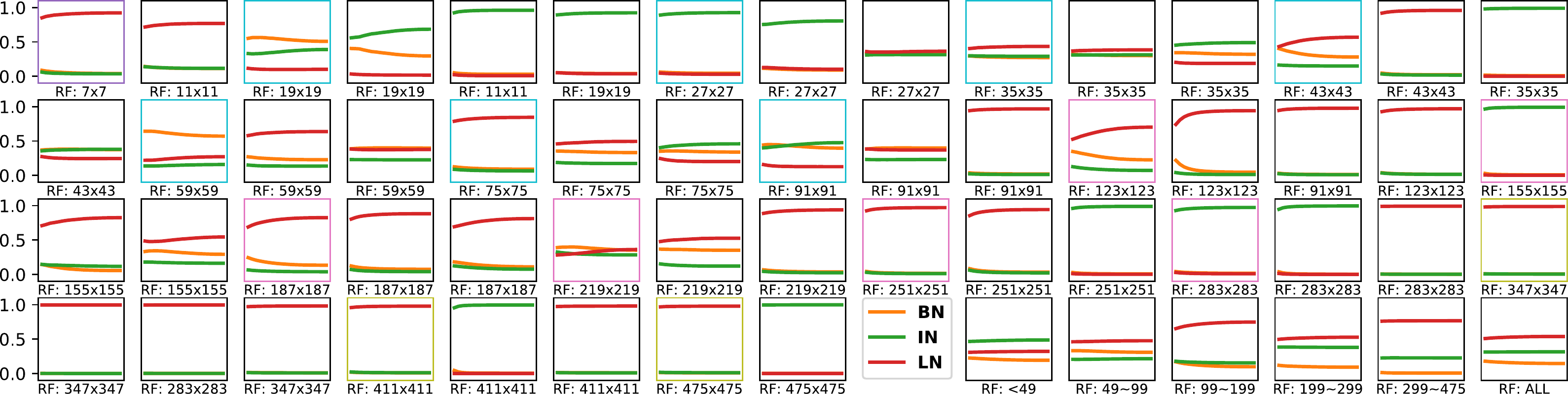}
		\vspace{-15pt}\caption{$\lambda_z^\sigma$ in \textbf{SN(8,4)} for ADE20K.}
	\end{subfigure}
~
\begin{subfigure}{0.95\textwidth}
\centering
\includegraphics[width=\textwidth,height=35mm]{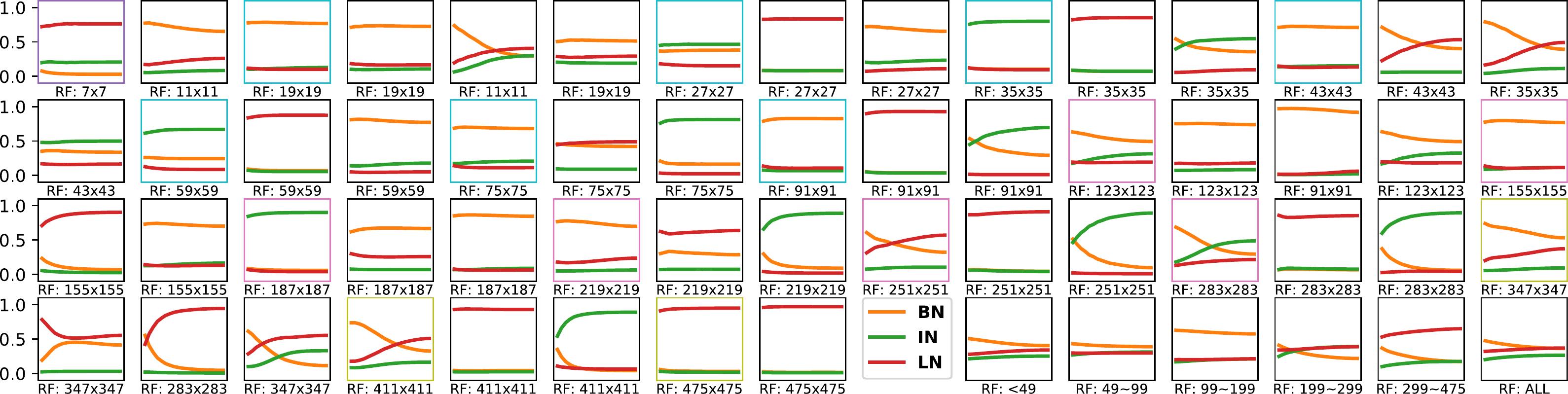}
		\vspace{-15pt}\caption{$\lambda_z^\mu$ in \textbf{SN(8,32)} for ADE20K.}
	\end{subfigure}
~
\begin{subfigure}{0.95\textwidth}
\centering
\includegraphics[width=\textwidth,height=35mm]{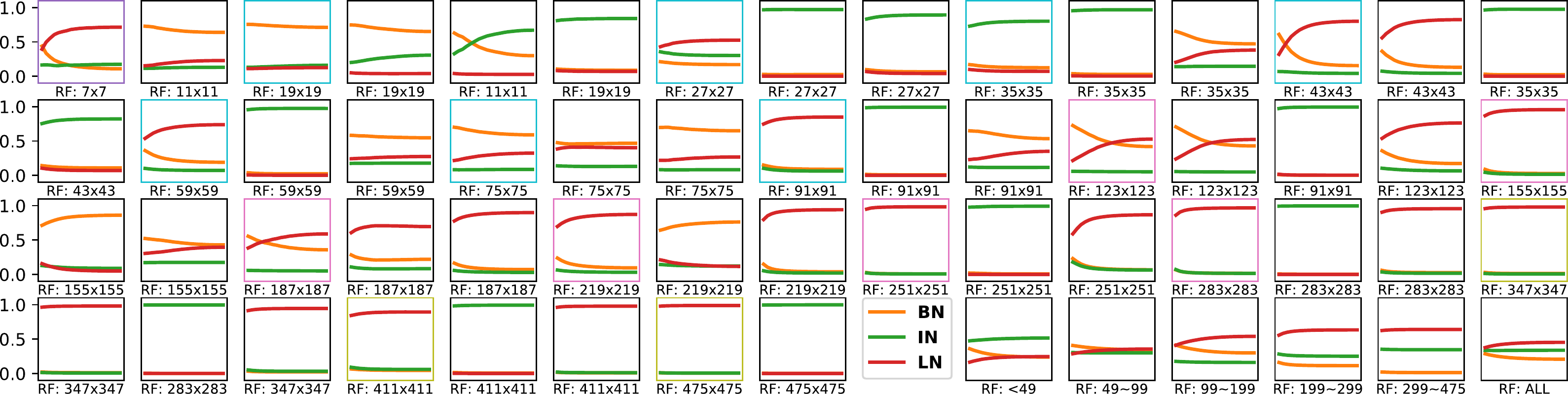}
		\vspace{-15pt}\caption{$\lambda_z^\sigma$ in \textbf{SN(8,32)} for ADE20K.}
	\end{subfigure}
\vspace{-5pt}
\caption{{Finetuning ResNet50+SN in ADE20K.
}}
\label{appfig:seg_training_ade}
\end{figure*}

\begin{figure*}[t]
%\vspace{-10pt}
%\hspace{-25pt}
\centering
	\begin{subfigure}{1\textwidth}
\centering
\includegraphics[width=\textwidth,height=22mm]{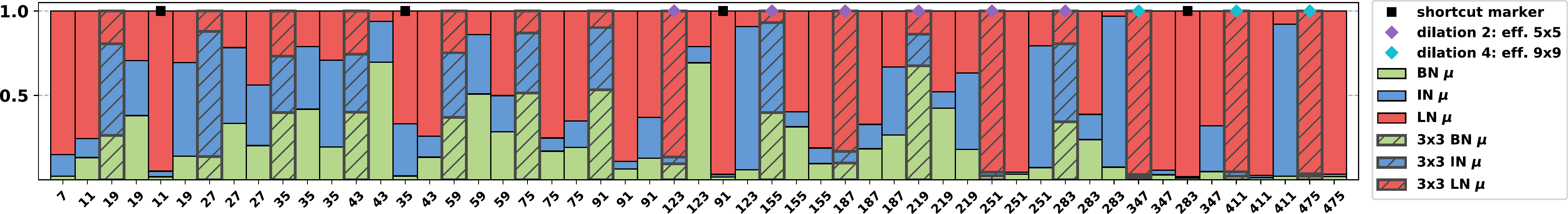}
		\vspace{-15pt}\caption{$\lambda_z^\mu$ for \textbf{SN(8,2)} in ADE20K.}
	\end{subfigure}
	~
	\begin{subfigure}{1\textwidth}
\centering
\includegraphics[width=\textwidth,height=22mm]{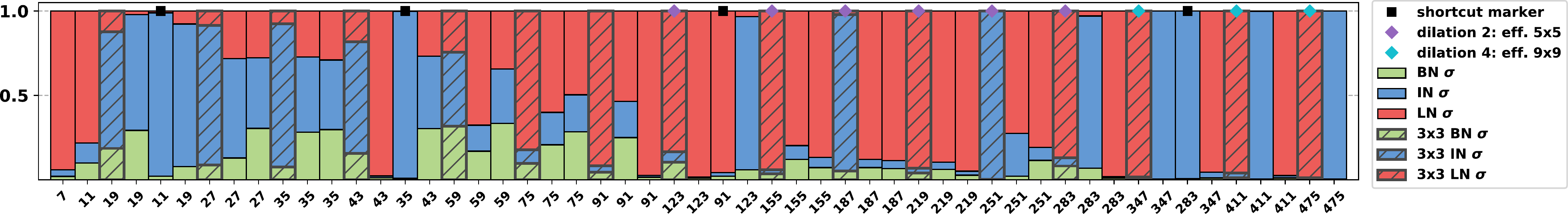}
		\vspace{-15pt}\caption{$\lambda_z^\sigma$ for \textbf{SN(8,2)} in ADE20K.}
	\end{subfigure}
~
\begin{subfigure}{1\textwidth}
\centering
\includegraphics[width=\textwidth,height=22mm]{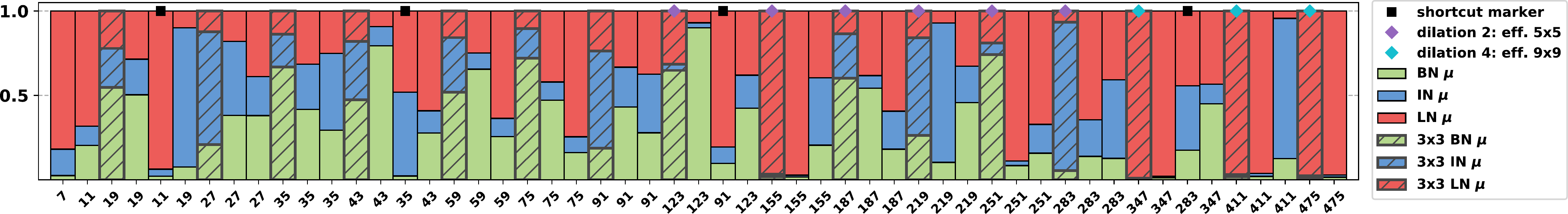}
		\vspace{-15pt}\caption{$\lambda_z^\mu$ for \textbf{SN(8,4)} in ADE20K.}
	\end{subfigure}
~
\begin{subfigure}{1\textwidth}
\centering
\includegraphics[width=\textwidth,height=22mm]{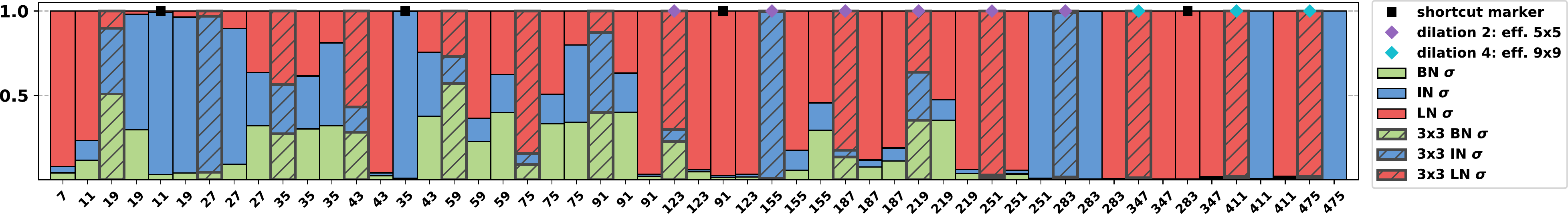}
		\vspace{-15pt}\caption{$\lambda_z^\sigma$ for \textbf{SN(8,4)} in ADE20K.}
	\end{subfigure}
~
\begin{subfigure}{1\textwidth}
\centering
\includegraphics[width=\textwidth,height=22mm]{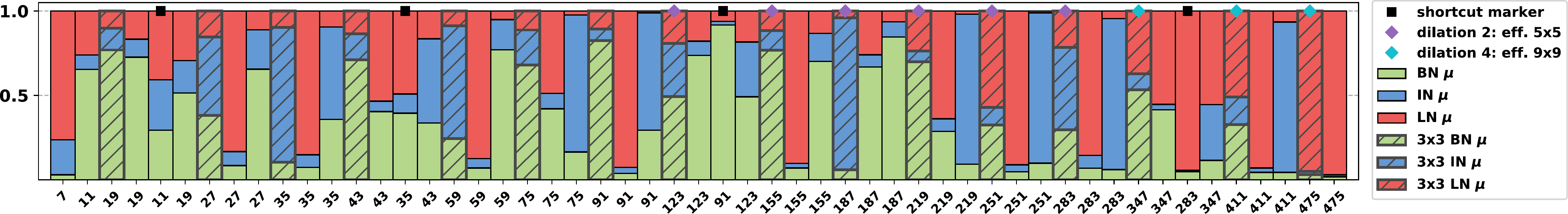}
		\vspace{-15pt}\caption{$\lambda_z^\mu$ for \textbf{SN(8,32)} in ADE20K.}
	\end{subfigure}
~
\begin{subfigure}{1\textwidth}
\centering
\includegraphics[width=\textwidth,height=22mm]{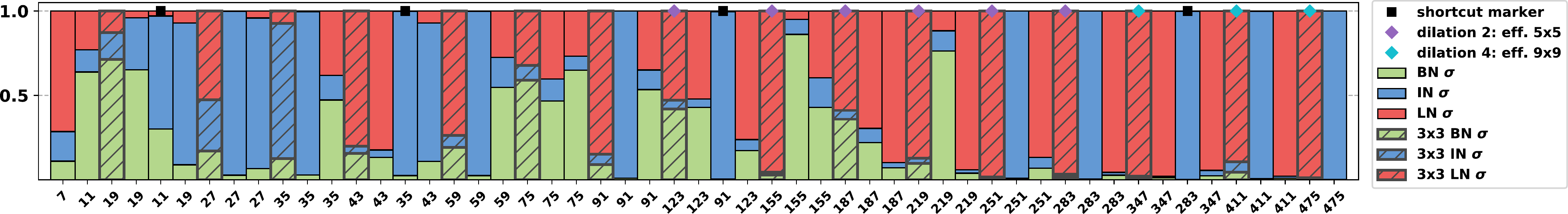}
		\vspace{-15pt}\caption{$\lambda_z^\sigma$ for \textbf{SN(8,32)} in ADE20K.}
	\end{subfigure}
\vspace{-5pt}
\caption{{$\lambda_z^\mu$ and $\lambda_z^\sigma$ when finetuning ResNet50+SN converged.
}}
\label{appfig:segm_converage_ade}
\end{figure*}

\end{document}